\documentclass[arxiv,article,accept,oneauthor,pdftex]{Definitions/mdpi} 

\firstpage{1} 
\pubvolume{12}
\issuenum{11}
\articlenumber{5506}
\pubyear{2022}
\copyrightyear{2022}
\datereceived{6 May 2022} 
\dateaccepted{27 May 2022} 
\datepublished{29 May 2022} 
\hreflink{https://doi.org/10.3390/\linebreak app12115506} 
\doinum{10.3390/app12115506}
\pdfoutput=1

\usepackage{tikz}
\usepackage[noend]{algorithmic}
\usepackage{algorithm}
\usepackage{amsmath}
\DeclareMathOperator*{\argmin}{argmin}

\Title{Cycle Mutation: Evolving Permutations via Cycle Induction}

\TitleCitation{Cycle Mutation: Evolving Permutations via Cycle Induction}

\Author{Vincent A. Cicirello \orcidA{}}

\AuthorNames{Vincent A. Cicirello}

\AuthorCitation{Cicirello, V.A.}

\address[1]{%
$^{1}$ \quad Stockton University, 101 Vera King Farris Dr, Galloway, NJ 08205; cicirelv@stockton.edu}

\corres{Correspondence: cicirelv@stockton.edu}

\abstract{Evolutionary algorithms solve problems by simulating the evolution of a 
population of candidate solutions. We focus on evolving permutations
for ordering problems like the traveling salesperson problem (TSP), as well as 
assignment problems like the quadratic assignment problem (QAP) and largest 
common subgraph (LCS). We propose cycle mutation, a new mutation operator
whose inspiration is the well known cycle crossover operator, and the concept 
of a permutation cycle. We use fitness landscape analysis to explore the 
problem characteristics for which cycle mutation works best. As a 
prerequisite, we develop new permutation distance measures: cycle distance, 
$k$-cycle distance, and cycle edit distance. The fitness landscape analysis 
predicts that cycle mutation is better suited for assignment and mapping problems 
than it is for ordering problems. We experimentally validate these findings 
showing cycle mutation's strengths on problems like QAP and LCS, and its 
limitations on problems like the TSP, while also showing that it is less prone 
to local optima than commonly used alternatives. We integrate cycle mutation 
into the open-source Chips-n-Salsa library, and the new distance metrics into 
the open-source JavaPermutationTools library.}



\begin{document}

\section{Introduction}\label{sec:intro}

In an Evolutionary Algorithm (EA), a problem is solved through the simulated evolution
of a population that evolves over many generations. There are many types of EA that mostly
differ in the types of problems they solve and in how solutions are represented. For example,
a Genetic Algorithm (GA)~\cite{Mitchell1998}, the original EA, usually represents a candidate 
solution to an optimization problem with a vector of bits. Evolution Strategies (ES)~\cite{Beyer2013}
focuses specifically on real-valued function optimization, utilizing a vector of floating-point values
to represent each candidate solution. Genetic Programming~\cite{Langdon2013}
is an approach to automatic inductive programming, and evolves a population of programs, each of which
is typically represented with a tree structure.

This paper focuses on EAs that encode solutions with permutations, often 
referred to as a Permutation-Based GA~\cite{Cuellar2021, Koohestani2020, Shabash2013, Kalita2019}, 
while others prefer the more general term EA~\cite{Shakya2017, Mironovich2021} to avoid confusion 
with a binary encoded GA. Solutions to some problems are more naturally represented
with a permutation than with another representation. The classic example is the Traveling Salesperson 
Problem (TSP), where a solution is a tour of the cities, and thus can be represented in a 
straightforward way as a permutation of indexes into a list of cities.

One challenge with designing a permutation EA is deciding which genetic operators to use.
This is less of an issue with the classic bit-vector GA or a real-valued ES because it is
possible to mutate bits independently of the rest of a bit-vector in a GA, and real-valued
alleles in an ES can likewise be mutated independently from the vector as a whole, such as 
with Gaussian mutation~\cite{Hinterding1995} or Cauchy mutation~\cite{Szu1987}. Within a 
permutation-based EA, mutation cannot change individual elements independent of the rest of the 
permutation. For example, in the permutation $[2, 4, 0, 3, 1]$ of the first five 
integers, we cannot mutate one value in isolation or it leads to an invalid permutation. 
Crossover cannot naively exchange parts of parents. For example, if one parent is as above, 
and the other is $[4, 2, 1, 0, 3]$, exchanging the second and third elements between the 
parents leads to two invalid permutations, $[2, 2, 1, 3, 1]$ and $[4, 4, 0, 0, 3]$.
Thus, for permutations, mutation and crossover must consider the overall structure to ensure
a valid encoding. 

As a consequence, many mutation and crossover operators exist
for permutations. The suitability of each depends upon
the characteristics of the permutation that most significantly impacts solution 
fitness for the problem at hand, such as absolute element positions, relative 
element positions, or element precedences~\cite{campos2005,cicirello2019}. 
Only relative positions matter to fitness of a TSP solution. For 
example, if cities $i$ and $j$ are adjacent in the permutation, then
the solution includes the edge $(i, j)$ regardless of where the pair of cities
appears. The absolute element positions are most important for other problems,
such as the Largest Common Subgraph (LCS). The LCS is an NP-Hard~\cite{Garey1979} 
optimization problem involving finding a one-to-one mapping between the vertex 
sets of a pair of graphs to maximize the number of edges of the common subgraph 
implied by the mapping. As a permutation problem, one holds the vertexes of one 
graph in a fixed order, and a permutation of the vertexes of the other graph 
represents a mapping. The absolute index of a vertex in the permutation
therefore corresponds to the vertex it is mapped to in the other graph.

Mutation for permutations is most often one of swap mutation, insertion mutation, 
reversal mutation, or scramble mutation~\cite{Eiben2015}. There are many permutation
crossover operators that focus on maintaining different characteristics of the parents,
including order crossover~\cite{ox}, non-wrapping order crossover~\cite{Cicirello2006}, 
uniform order-based crossover~\cite{uobx}, partially matched crossover~\cite{pmx}, 
uniform partially matched crossover~\cite{Cicirello2000}, precedence preservative 
crossover~\cite{ppx}, edge assembly crossover~\cite{NagataEAX,eax}, and Cycle 
Crossover (CX)~\cite{cx}.

The central aim of this paper is development of a mutation operator for
permutations that: (a) is characterized by small random perturbations on average,
(b) has a large neighborhood size, and (c) is tunable. These properties
enable focused search with improved handling of local optima. No such mutation 
operators currently exist for permutations. For bit-vectors, the standard 
bit-flip mutation satisfies all of these properties. For example, the bit-flip
mutation rate $M$ is usually set to a low value for a small average number of
bits flipped, but can be tuned for problems where greater mutation is beneficial,
and as long as $M$ is non-zero the neighborhood includes all other bit-vectors.
Similarly, Gaussian mutation in an ES satisfies all of these properties,
with a tunable parameter, the standard deviation of the Gaussian. All
existing permutation mutation operators have a fixed neighborhood size, which in
most cases is small, that depends only on permutation length.

To achieve this aim, we present a new mutation operator called 
{\em cycle mutation}, which relies on the concept of a permutation cycle
and is inspired by CX. Rather than operating on two parents as CX 
does, cycle mutation instead mutates a single member of the population. 
Thus, it is also applicable to non-population 
metaheuristics such as Simulated Annealing 
(SA)~\cite{Delahaye2019,Laarhoven1987,Kirkpatrick1983}. We develop two 
variations of cycle mutation, $\mathrm{Cycle}(\mathit{kmax})$ and $\mathrm{Cycle}(\alpha)$,  
offering two ways of addressing locally optimal solutions. 

To formally demonstrate that cycle mutation achieves the target properties 
of large neighborhood but small average changes, we conduct a fitness 
landscape analysis of cycle mutation, and other permutation mutation 
operators for three NP-Hard optimization problems~\cite{Garey1979}, 
the TSP, the LCS, and the Quadratic Assignment Problem (QAP). The 
fitness landscape analysis predicts that cycle mutation likely performs 
well for assignment and mapping problems, such as LCS and QAP, where 
absolute positions directly affect fitness, but that it may be less 
well-suited to relative ordering problems like the TSP. We use 
Fitness Distance Correlation (FDC)~\cite{fdc}, 
which requires a measure of distance between solutions that corresponds to 
the operator under analysis. Appropriate measures of distance exist for the 
operators to which we compare. However, no existing permutation distance 
functions are suitable for the new cycle mutation. Therefore, we introduce 
three new permutation distance measures: {\em cycle distance}, 
{\em $k$-cycle distance}, and {\em cycle edit distance}.

To validate the new cycle mutation, we experiment with the LCS, QAP, and TSP,
comparing cycle mutation with commonly used permutation mutation operators 
within a $(1+1)$-EA as well as within SA. The results support the predictions 
of the fitness landscape analysis, demonstrating the efficacy of cycle mutation for 
assignment problems like LCS and QAP, while also showing that cycle 
mutation is inferior to alternatives for the TSP where relative element 
positions or more important than absolute locations. The experiments also show 
that the $\mathrm{Cycle}(\alpha)$ variation is especially effective at escaping local optima.

A secondary aim of this research is to enable reproducibility~\cite{NAP25303},
as well as to advance the state of practice. Therefore, we integrate our 
Java implementation of cycle mutation into the open source Chips-n-Salsa 
library~\cite{Cicirello2020joss}, and cycle distance, $k$-cycle distance, 
and cycle edit distance into the open source JavaPermutationTools 
library~\cite{cicirello2018}. Chips-n-Salsa is a library for stochastic
and adaptive local search as well as EAs. JavaPermutationTools is a library 
for computation on permutations and sequences, with a focus on measures of distance. 
We also disseminate the source code for the fitness landscape analysis
and experiments, as well as the raw and processed experiment data,  
on GitHub (\url{https://github.com/cicirello/cycle-mutation-experiments}). 

We begin by introducing necessary background in Section~\ref{sec:background}. We 
proceed with our methods in Section~\ref{sec:methods}, including deriving the new 
cycle mutation and distance metrics, as well as performing the fitness landscape 
analysis. Results are presented in Section~\ref{sec:experiments}. We wrap up with 
a discussion and conclusions in Section~\ref{sec:conclusions} including discussing
insights into the situations where the new cycle mutation is likely to excel.

\section{Background}\label{sec:background}

This section provides background on common mutation operators for 
permutations (Section~\ref{sec:mutationops}); permutation
cycles (Section~\ref{sec:cycles}), which is the theoretical basis for the new cycle
mutation; the CX operator (Section~\ref{sec:cx}) on which the new cycle mutation is based;
and NP-Hard combinatorial optimization problems (Section~\ref{sec:problems}) used later in 
this article.

\subsection{Permutation Mutation Operators}\label{sec:mutationops}

We later compare cycle mutation to the most common permutation mutation operators,
including the following.
Swap picks two different elements uniformly at random and exchanges their locations
within the permutation. All other elements remain in their current positions.
Insertion picks an element uniformly at random, removes it from the permutation,
and reinserts it at a different position chosen uniformly at random. This has the 
effect of shifting all elements between the removal and insertion points.
Reversal (also known as inversion) reverses the order of a subsequence
of the permutation, where the end points are chosen uniformly at random.
Scramble (also known as shuffle) randomizes the order of a subsequence
of the permutation, where the end points of the subsequence are chosen uniformly 
at random. Scramble is the most disruptive of these operators.

Prior fitness landscape analyses (e.g.,~\cite{Cicirello2016}) show that swap 
strongly dominates when absolute 
positions are most important to fitness; reversal is best for relative positions 
with undirected edges, followed by insertion and swap, but reversal 
performs poorly with directed edges; and insertion is best when 
element precedences most greatly affect fitness.

\subsection{Permutation Cycles}\label{sec:cycles}

Cycle mutation relies upon the concept of a permutation cycle~\cite{KnuthVolume1}. Align
two permutations such that corresponding positions are vertically adjacent, such as:
\begin{align}\label{eq:pair}
\begin{split}
p_1 &= [0, 1, 2, 3, 4, 5, 6, 7, 8, 9] \\
p_2 &= [2, 3, 0, 5, 6, 7, 8, 9, 4, 1] .
\end{split}
\end{align}
Consider a directed graph with one vertex for each element. In this example, the
hypothetical graph has 10 vertexes. Corresponding positions define directed edges.
For example, $p_1$ and $p_2$ have 0 and 2 at the beginning, respectively, which
implies an edge from vertex 0 to vertex 2. Thus, the directed edges of the graph 
induced by $p_1$ and $p_2$ are:
\begin{equation}
\{ (0,2), (1,3), (2,0), (3,5), (4,6), (5,7), (6,8), (7,9), (8,4), (9,1) \} .
\end{equation} 
A permutation cycle is a cycle in this graph. Thus, in this example, there are three permutation
cycles, consisting of the following sets of vertexes:
\begin{equation}\label{eq:cycles}
\{ 0, 2 \}, \{ 1, 3, 5, 7, 9 \}, \{ 4, 6, 8 \}.
\end{equation}

\subsection{Cycle Crossover (CX)}\label{sec:cx}

The CX~\cite{cx} operator creates two children from two parents as follows. It first
selects an index into one parent uniformly at random. It computes
the permutation cycle for the pair of parents that includes the chosen element.
Child $c_1$ gets the positions of the elements that are in the cycle from $p_2$ and the positions
of the other elements from $p_1$. Likewise, child $c_2$ gets the positions 
of the elements that are in the cycle from $p_1$ and the positions of the others 
from $p_2$. The runtime to apply CX is $\Theta(n)$ where
$n$ is the permutation length.

Consider an example where the parents are the permutations of Equation~\ref{eq:pair}, which consist
of three permutation cycles (see Equation~\ref{eq:cycles}). The result of 
CX depends upon the random starting element. If element 0 or 2 begins the cycle, then 
the children are:
\begin{align}
\begin{split}
c_1 &= [2, 1, 0, 3, 4, 5, 6, 7, 8, 9] \\
c_2 &= [0, 3, 2, 5, 6, 7, 8, 9, 4, 1] .
\end{split}
\end{align}
If one of the elements $\{ 1, 3, 5, 7, 9 \}$ begins the cycle, then the children are:
\begin{align}
\begin{split}
c_1 &= [0, 3, 2, 5, 4, 7, 6, 9, 8, 1] \\
c_2 &= [2, 1, 0, 3, 6, 5, 8, 7, 4, 9] .
\end{split}
\end{align}
Otherwise, if one of the elements $\{ 4, 6, 8 \}$ begins the cycle, then the children are:
\begin{align}
\begin{split}
c_1 &= [0, 1, 2, 3, 6, 5, 8, 7, 4, 9] \\
c_2 &= [2, 3, 0, 5, 4, 7, 6, 9, 8, 1] .
\end{split}
\end{align}
Since the starting element is chosen uniformly at random, larger cycles are chosen
with greater probability. In this example, the probability of generating the first pair of
children above is 0.2, while the probability of generating the second pair of children is 0.5,
and the probability of generating the last pair of children is 0.3.

One characteristic of CX is that every element of each child gets its absolute position within
the permutation from one or the other of the two parents. In this way, it is particularly well-suited
to permutation problems where absolute position has greatest effect on fitness, since the children
are inheriting absolute element position from the parents.

\subsection{Test Problems}\label{sec:problems}

We now provide background on the TSP, QAP, and LCS, which
are NP-Hard combinatorial optimization problems used in the
fitness landscape analysis and experiments.

\subsubsection{TSP} 
In the TSP, a salesperson must complete a tour of
$n$ cities to minimize total cost, usually distance traveled. 
The cities are vertexes of a completely connected graph. A tour
is a simple cycle that includes all $n$ vertexes. The NP-Complete decision 
variant of the TSP asks whether a tour exists with cost at most $C$; and
the NP-Hard optimization problem seeks the minimum cost tour~\cite{Garey1979}.  
The TSP is widely studied and is perhaps the most common 
combinatorial optimization problem in experimental studies. It has been used in
machine learning~\cite{Mele2021,Mele2021-2}, ant colony 
optimization~\cite{Wang2021,Dinh2021}, GA~\cite{Varadarajan2021,Nagata2020,cx,pmx}, other forms of 
EA~\cite{Ibada2021,NagataEAX,eax}, and other  
metaheuristics~\cite{DellAmico2021,Gao2021,Tong2022,Qamar2021,RicoGarcia2021}. There 
are variations of the problem, such as the Asymmetric TSP (ATSP), where the cost of using an 
edge differs depending upon the direction of travel along the edge~\cite{An2021,Svensson2020}.
Some variations include problem domain characteristics, such as in
delivery route planning~\cite{Tong2022} and wireless sensor networks~\cite{Tsilomitrou2022}.

\subsubsection{QAP} 
The formal definition of the QAP is as follows. We are given an $n$ by $n$ cost matrix $C$,
and an $m$ by $m$ distance matrix $D$, such that $m \geq n$. The NP-Complete decision version of the 
problem~\cite{Garey1979} asks the question, for a given bound $B$, does there exist a one-to-one function
$f:\{0, 1, \ldots, n-1\}\rightarrow\{0, 1, \ldots, m-1\}$, such that:
\begin{equation}
\sum_{i=0}^{n-1} \sum_{\substack{j=0 \\ j \neq i}}^{n-1} C_{i,j} D_{f(i),f(j)} \leq B ?
\end{equation}
The NP-Hard optimization version of the QAP is to find the one-to-one function
$f:\{0, 1, \ldots, n-1\}\rightarrow\{0, 1, \ldots, m-1\}$ that minimizes:
\begin{equation}
\sum_{i=0}^{n-1} \sum_{\substack{j=0 \\ j \neq i}}^{n-1} C_{i,j} D_{f(i),f(j)} .
\end{equation}
Most authors consider the case when $n=m$. This problem is
naturally represented with permutations. When $n=m$, a solution (i.e., the one-to-one
function $f$) is simply represented as a permutation of the integers $\{0, 1, \ldots, n-1 \}$.
In the more general case, a solution is the first $n$ integers in a permutation
of the integers $\{0, 1, \ldots, m-1 \}$. The QAP is an especially
challenging NP-Hard optimization problem, where you most often find experimental studies utilizing
what seems like rather small instances (e.g., $n < 50$). A wide variety of EA, metaheuristics, and heuristic
approaches have been proposed for the 
QAP~\cite{He2018,Beham2017,Benavides2021,Baioletti2019,Novaes2019,Thomson2017,Irurozki2018}.

\subsubsection{LCS} 
The LCS problem~\cite{Garey1979} is closely related to the subgraph isomorphism problem. In the LCS problem,
we are given graphs $G_1 = (V_1, E_1)$ and $G_2 = (V_2, E_2)$, with vertex sets $V_1$ and $V_2$, and edge sets
$E_1$ and $E_2$. In the optimization variant of the problem, we must find the 
graph $G_3 = (V_3, E_3)$, such that $G_3$ is isomorphic to a subgraph of $G_1$
and $G_3$ is isomorphic to a subgraph of $G_2$, and such that the cardinality of edge set $|E_3|$ is maximized.
This problem is NP-Hard. The NP-Complete decision variant of the problem asks whether there exists
such a graph $G_3$ such that $|E_3| \geq K$ for some threshold $K$.
 
If $|V_1| = |V_2|$, then a solution is represented by
a permutation of $\{ 0, 1, \ldots, |V_1| - 1\}$, and more generally
the solution is represented by the first $\min(|V_1|, |V_2|)$ integers of a permutation of 
$\{ 0, 1, \ldots, \max(|V_1|, |V_2|) - 1\}$. Without loss of generality, assume that 
$|V_1| \leq |V_2|$. For permutation $p$ of integers $\{ 0, 1, \ldots, |V_2| - 1\}$, 
The number of edges $|E_3|$ in the common subgraph implied by such a permutation 
is then computed as:
\begin{equation}
|E_3| = \sum_{(u,v) \in E_1} \begin{cases}
1 & \text{if $(p(u),p(v)) \in E_2$} \\
0 & \text{if $(p(u),p(v)) \notin E_2$} ,
\end{cases}
\end{equation}
where $p(u)$ means the element in position $u$ of permutation $p$. This is most efficiently 
implemented if the smaller graph, $G_1$, is represented with adjacency lists or by a simple
list of ordered pairs as the edge set; and if the larger graph $G_2$ is represented with
an adjacency matrix to enable constant time checks for existence of edges. 

Prior approaches to the LCS problem include GA~\cite{Cicirello2000}, hill-climbing~\cite{cicirello2002}, 
and heuristic approaches~\cite{Chen2020,Stoichev2009,Zeller2002}.
There are applications of the LCS problem in computer-aided engineering~\cite{cicirello2002},
software engineering~\cite{Zeller2002}, protein molecule comparisons~\cite{Stoichev2009},
integrated circuit design~\cite{Chen2020,Wong2005}, natural language processing~\cite{Jordan2006},
cybersecurity~\cite{Puodzius2021}, among others.

\section{Methods}\label{sec:methods}

To achieve the objective of a tunable mutation operator with a large neighborhood
but small average changes, we derive two forms of the new cycle mutation in 
Section~\ref{sec:mutation}. 

To gain an understanding of the topology of fitness landscapes associated 
with cycle mutation, we use a variety of fitness landscape analysis tools. 
This requires measures of permutation distance corresponding to the mutation 
operators. Ideally, these should be edit distances where the mutation operator 
is the edit operation. Edit distance is defined as the minimum cost of the 
edit operations necessary to transform one structure into the other, and 
originated within the context of string distance~\cite{wagner74,levenshtein1966}. 
There are many permutation distance measures available in the 
literature~\cite{cicirello2019,cicirello2018,sorensen07,schiavinotto2007,campos2005,ronald1998,ronald1997}, 
including several edit distances. However, none of these are suitable for 
characterizing the distance between permutations within the context of cycle mutation.
Therefore, in Section~\ref{sec:distance}, we derive new measures of permutation 
distance: {\em cycle distance}, {\em cycle edit distance}, and {\em $k$-cycle distance}.

We then proceed with the fitness landscape analysis in Section~\ref{sec:landscapes}.
We derive the diameter of the fitness landscapes for both forms of cycle 
mutation, as well as for other common permutation mutation operators. The diameter of a 
fitness landscape is the distance between the two furthest points, where distance  
is the minimum number of applications of the operators to transform one point 
to the other. Thus, diameter directly relates to neighborhood size, where larger neighborhood 
corresponds to smaller diameter, providing a means to quantify our objective of a mutation
with large neighborhood.
The fitness landscape analysis then utilizes FDC, which is the Pearson 
correlation coefficient between the fitness of solutions, and the distance 
to the nearest optimal solution~\cite{fdc}. The
FDC analysis uses the TSP, LCS, and QAP problems, so that we can explore the behavior
of the cycle mutation operator on a variety of permutation problems.
To directly address the objective of mutation with small average changes, 
the fitness landscape analysis also uses search landscape calculus~\cite{Cicirello2016}, 
which examines the average local rate of fitness change.

\subsection{Cycle Mutation}\label{sec:mutation}

We present two variations of cycle mutation. Both forms mutate permutation $p$ by inducing a 
permutation cycle of length $k$. The primary difference between the two versions is in how
$k$ is chosen. We provide notation and operations shared by the two variations in 
Section~\ref{sec:shared}, followed by the presentations of the two versions, $\mathrm{Cycle}(\mathit{kmax})$ in 
Section~\ref{sec:kmax} and $\mathrm{Cycle}(\alpha)$ in Section~\ref{sec:alpha}, and finally
a comparison of the asymptotic runtime of the new cycle mutation with commonly
used mutation operators in Section~\ref{sec:runtime}.

\subsubsection{Shared Notation and Operations}\label{sec:shared}

Let $p$ be a permutation of length $n$,
such that $p(i)$ is the element at index $i$. Assume
that indexes into $p$ are 0-based (i.e., valid indexes are $\{0, 1, \ldots, n-1\}$),
as are array indexes.
Algorithm~\ref{alg:createCycle} shows pseudocode for the core operation of both
variations of cycle mutation. Namely, it induces a cycle in the given permutation $p$ 
from an array of indexes into $p$.

\begin{algorithm}[H]
\caption{$\mathrm{CreateCycle}(p, \mathrm{indexes})$}
\algsetup{indent=2em}
\begin{algorithmic}[1]
\STATE $\mathrm{temp} \leftarrow p(\mathrm{indexes}[0])$
\FOR{$i = 1$ \TO $n-1$}
	\STATE $p(\mathrm{indexes}[i-1]) \leftarrow p(\mathrm{indexes}[i])$
\ENDFOR
\STATE $p(\mathrm{indexes}[n-1]) \leftarrow \mathrm{temp}$
\end{algorithmic}
\label{alg:createCycle}
\end{algorithm}

As an example of its behavior, consider the following permutation:
\begin{equation}
p = [2, 6, 0, 5, 3, 8, 7, 9, 4, 1] .
\end{equation}
Now consider the following array of indexes into $p$:
\begin{equation}
\mathrm{indexes} = [ 3, 7, 1, 4] .
\end{equation}
Executing $\mathrm{CreateCycle}(p, \mathrm{indexes})$ will
produce the permutation $p'$ such that $p'(3) = p(7)$,
$p'(7) = p(1)$, $p'(1) = p(4)$, $p'(4) = p(3)$, resulting in
the following:
\begin{equation}
p' = [2, 3, 0, 9, 5, 8, 7, 6, 4, 1] .
\end{equation}
The runtime of $\mathrm{CreateCycle}$ is linear in the induced cycle length.

One of the steps of cycle mutation requires sampling
$k$ random indexes into permutation $p$ without replacement. Algorithm~\ref{alg:sample}
shows our sampling approach, which uses one of three algorithms depending on the 
value of $k$ relative to the permutation length $n$. 

\begin{algorithm}[H]
\caption{$\mathrm{Sample}(n, k)$}
\algsetup{indent=2em}
\begin{algorithmic}[1]
\IF{$k \geq \frac{n}{2}$}
\RETURN $\mathrm{ReservoirSample}(n, k)$
\ELSIF{$k \geq \sqrt{n}$}
\RETURN $\mathrm{PoolSample}(n, k)$
\ELSE
\RETURN $\mathrm{InsertionSample}(n, k)$
\ENDIF
\end{algorithmic}
\label{alg:sample}
\end{algorithm}

When $k \geq \frac{n}{2}$, we use Vitter's reservoir sampling algorithm~\cite{Vitter1985} (line 
2 of Algorithm~\ref{alg:sample}), which has a runtime of $O(n)$ and utilizes $O(n-k)$ random numbers. 
When $\sqrt{n} \leq k < \frac{n}{2}$, we use Ernvall and Nevalainen's sampling 
algorithm~\cite{Ernvall1982}, which we refer to as $\mathrm{PoolSample}$ in line 4 of 
Algorithm~\ref{alg:sample}, and which also has a runtime of $O(n)$, but requires $O(k)$
random numbers. Asymptotic runtime of both of these options is $O(n)$, but since
random number generation is a costly operation with very significant impact on the runtime of
an EA~\cite{cicirello2018-2}, our approach chooses the sampling algorithm that requires fewer 
random numbers. 

When $k < \sqrt{n}$, we use what we believe is a brand new sampling algorithm:
Insertion Sampling (Algorithm~\ref{alg:sample}, line 6). Insertion sampling's
runtime is $O(k^2)$, and requires $O(k)$ random numbers. Since runtime increases
quadratically in $k$, insertion sampling is only a good choice when $k$ is very small
relative to $n$. Pseudocode for insertion sampling is in Algorithm~\ref{alg:insertionSample}.
To ensure a duplicate-free result, it maintains a sorted list of
the integers selected thus far, and inserts into that list in a way similar to insertion sort.
The $\mathrm{Rand}(a,b)$ in Algorithm~\ref{alg:insertionSample} is a uniform random 
variable over the interval $[a,b]$, inclusive.

\begin{algorithm}[H]
\caption{$\mathrm{InsertionSample}(n, k)$}
\algsetup{indent=2em}
\begin{algorithmic}[1]
\STATE $\mathrm{result} \leftarrow \text{ a new array of length } k$
\FOR{$i = 0$ \TO $k-1$}
	\STATE $v \leftarrow \mathrm{Rand}(0, n-i-1)$
	\STATE $j \leftarrow k - i$
	\WHILE{$j < k$ \AND $v \geq \mathrm{result}[j]$}
		\STATE $v \leftarrow v + 1$
		\STATE $\mathrm{result}[j-1] \leftarrow \mathrm{result}[j]$
		\STATE $j \leftarrow j + 1$
	\ENDWHILE
	\STATE $\mathrm{result}[j-1] \leftarrow v$
\ENDFOR 
\RETURN $\mathrm{result}$
\end{algorithmic}
\label{alg:insertionSample}
\end{algorithm}

The composite sampling algorithm of Algorithm~\ref{alg:sample}
runs in $O(\min(n,k^2))$ time and requires $O(\min(k,n-k))$ random numbers.
We integrated this composite sampling algorithm, the new insertion sampling
algorithm, as well as our implementations of reservoir sampling and pool sampling into an
open source Java library $\rho\mu$ (\url{https://github.com/cicirello/rho-mu}), 
independent of the application to EA of this article.

\subsubsection{$\mathrm{Cycle}(\mathit{kmax})$}\label{sec:kmax}

In the first version of cycle mutation (Algorithm~\ref{alg:cycleMutationkmax}), 
called $\mathrm{Cycle}(\mathit{kmax})$, the induced cycle length $k$ is 
uniformly random from the interval $[2, \mathit{kmax}]$, such that 
$\mathit{kmax} \geq 2$. 

\begin{algorithm}[H]
\caption{$\mathrm{CycleMutation}(p, \mathit{kmax})$}
\algsetup{indent=2em}
\begin{algorithmic}[1]
\STATE $k \leftarrow \mathrm{Rand}(2,\mathit{kmax})$
\STATE $\mathrm{indexes} \leftarrow \mathrm{Sample}(n, k)$
\STATE $\mathrm{Shuffle}(\mathrm{indexes})$
\STATE $\mathrm{CreateCycle}(p, \mathrm{indexes})$
\end{algorithmic}
\label{alg:cycleMutationkmax}
\end{algorithm}

It first selects the cycle length $k$ in $O(1)$time. Next, $k$ indexes are 
sampled uniformly at random without replacement (line 2) with a call to $\mathrm{Sample}(n, k)$ 
of Algorithm~\ref{alg:sample}. Since $k \leq \mathit{kmax}$, the worst case runtime of that step 
is $O(\min(n,\mathit{kmax}^2))$. This array of indexes is randomized (line 3), 
with a worst case cost of $O(\mathit{kmax})$. Finally, a 
cycle is induced from the randomized list of $k$ indexes with a 
call to the $\mathrm{CreateCycle}$ of Algorithm~\ref{alg:createCycle} in line 4, which also 
costs $O(\mathit{kmax})$ time in the worst case. Thus, the worst case runtime is 
$O(\min(n,\mathit{kmax}^2))$, since the most costly step is sampling the
indexes. The average case is also $O(\min(n,\mathit{kmax}^2))$ since the
average cycle length $\bar{k} = \frac{\mathit{kmax} + 2}{2}$ is proportional to $\mathit{kmax}$.  

$\mathrm{Cycle}(\mathit{kmax})$ limits the induced permutation cycle to a predetermined
maximum length, with all cycle lengths up to $\mathit{kmax}$ equally likely.
This enables tuning the size of the local neighborhood, 
with lower $\mathit{kmax}$ leading to a smaller neighborhood and higher $\mathit{kmax}$ 
creating a larger neighborhood. Thus, increasing $\mathit{kmax}$ may lead to fewer
local optima in the fitness landscape, but would also lead to a more disruptive and less
focused search.

\subsubsection{$\mathrm{Cycle}(\alpha)$}\label{sec:alpha}

The second version of cycle mutation, called $\mathrm{Cycle}(\alpha)$,
maximizes the size of the local neighborhood while retaining the local focus
of a small cycle length. $\mathrm{Cycle}(\alpha)$ allows any possible cycle length
$k \in [2, n]$, but chooses a smaller cycle length with higher probability than a greater
cycle length. Specifically, the probability of choosing cycle length $k$ is proportional
to $\alpha^{k-2}$. Thus, the probability $P(k)$ of choosing cycle length $k$ is:
\begin{equation}\label{eq:P}
P(k) = \frac{\alpha^{k-2}}{\sum_{i=2}^{n} \alpha^{i-2}} = \frac{\alpha^{k-2} (1 - \alpha)}{1 - \alpha^{n-1}} .
\end{equation}
The $\alpha$ is a parameter of the operator such that $0 < \alpha < 1$. The nearer $\alpha$ is to 0, the
more probabilistic weight is placed upon lower values of $k$ relative to higher values.

From this, we can derive a mathematical transformation from uniform random number 
$U \in [0.0, 1.0)$ to corresponding cycle length $k$. Choose $k$ according to the following:
\begin{equation}
k = \begin{cases}
	2 & \text{if $0 \leq U < P(2)$} \\
	j & \text{if $\sum_{i=2}^{j-1} P(i) \leq U < \sum_{i=2}^{j} P(i)$} .
	\end{cases}
\end{equation}
This is equivalent to:
\begin{equation}
k = \argmin_{j \in [2,n]} \left\{ \sum_{i=2}^{j} P(i) \right\} \text{ subject to } \left\{ \sum_{i=2}^{j} P(i) > U \right\} .
\end{equation}
Substituting Equation~\ref{eq:P} into the constraint and simplifying arrives at:
\begin{equation}
k = \argmin_{j \in [2,n]} \left\{ \sum_{i=2}^{j} P(i) \right\}  \text{ subject to } \left\{ \frac{1 - \alpha^{j-1}}{1 - \alpha^{n-1}} > U \right\} . 
\end{equation}
Solve the constraint for $j$ to derive:
\begin{equation}
k = \argmin_{j \in [2,n]} \left\{ \sum_{i=2}^{j} P(i) \right\}  \text{ subject to } \left\{ j > \frac{\log(1 - (1-\alpha^{n-1})U)}{\log(\alpha)} + 1 \right\} . 
\end{equation}
Since the summation inside the $\argmin$ increases as $j$ increases, this is equivalent to:
\begin{equation}
k = \min_{j \in [2,n]} \left\{ j \right\} \text{ subject to } \left\{ j > \frac{\log(1 - (1-\alpha^{n-1})U)}{\log(\alpha)} + 1 \right\} . 
\end{equation}
Finally, we compute $k$ from $U$ via:
\begin{equation}
k = 2 + \left\lfloor \frac{\log(1 - (1-\alpha^{n-1})U)}{\log(\alpha)} \right\rfloor .
\end{equation}
Since $\alpha$ is a parameter that does not change during the run, and since the permutation length $n$ is likewise fixed
based on the problem instance we are solving, the $(1-\alpha^{n-1})$ and the $\log(\alpha)$ are constants that can be
computed a single time at the start of the run. 

Algorithm~\ref{alg:cycleMutationAlpha} shows pseudocode of $\mathrm{Cycle}(\alpha)$.
The worst case runtime is $O(n)$, which occurs when
the random $k$ equals $n$, leading lines 3--4 to cost $O(n)$. 
The worst case is a rare occurrence. Since mutation is applied a very large number of times
during an EA, it is more meaningful to examine the average runtime of mutation.
To determine the average runtime, we must first compute the expected cycle length $E[k]$ as follows:
\begin{equation}\begin{aligned}
E[k] &= \sum_{k=2}^{n} k P(k) \\
	&= \frac{(1 - \alpha)}{1 - \alpha^{n-1}} \sum_{k=2}^{n} k \alpha^{k-2} \\
	&= \frac{2 - \alpha + n \alpha^{n} - n \alpha^{n-1} - \alpha^{n-1}}{(1 - \alpha)(1 - \alpha^{n-1})} \\
	&\leq  \lim_{n \to \infty} \frac{2 - \alpha + n \alpha^{n} - n \alpha^{n-1} - \alpha^{n-1}}{(1 - \alpha)(1 - \alpha^{n-1})} \\
	&= \frac{2 - \alpha}{1 - \alpha} .
\end{aligned}
\end{equation}
The expected cycle lengths for $\mathrm{Cycle}(0.25)$, $\mathrm{Cycle}(0.5)$, and 
$\mathrm{Cycle}(0.75)$ are $E[k] \leq 2\frac{1}{3}$, $E[k] \leq 3$, and $E[k] \leq 5$, respectively.
The average runtime of $\mathrm{Cycle}(\alpha)$ is therefore
$O(\min(n,(\frac{2 - \alpha}{1 - \alpha})^2))$ due to the call to $\mathrm{Sample}(n, k)$ in line 2 
of Algorithm~\ref{alg:cycleMutationAlpha}.

\begin{algorithm}[H]
\caption{$\mathrm{CycleMutation}(p, \alpha)$}
\algsetup{indent=2em}
\begin{algorithmic}[1]
\STATE $k \leftarrow 2 + \left\lfloor \frac{\log(1 - (1-\alpha^{n-1})U)}{\log(\alpha)} \right\rfloor$
\STATE $\mathrm{indexes} \leftarrow \mathrm{Sample}(n, k)$
\STATE $\mathrm{Shuffle}(\mathrm{indexes})$
\STATE $\mathrm{CreateCycle}(p, \mathrm{indexes})$
\end{algorithmic}
\label{alg:cycleMutationAlpha}
\end{algorithm}

\subsubsection{Asymptotic Runtime Summary}\label{sec:runtime}

Table~\ref{tab:runtime} summarizes the asymptotic runtime of cycle mutation and common
mutation operators. Swap mutation is a constant time operation, while the worst 
case runtime of insertion, reversal, scramble, and $\mathrm{Cycle}(\alpha)$ is linear in the permutation 
length $n$. The worst case for $\mathrm{Cycle}(\mathit{kmax})$ is between these extremes, 
depending upon $\mathit{kmax}$.

Since mutation is computed a very large number of times during an EA, average runtime
is more meaningful. The average runtime of insertion, reversal, and scramble is $O(n)$.
The average runtime of $\mathrm{Cycle}(\alpha)$ depends upon $\alpha$. However, other than values 
of $\alpha$ very near 1.0, the average runtime is essentially a constant.
Although the runtime of swap is constant, and that of cycle
mutation is very nearly constant depending on $\alpha$ or $\mathit{kmax}$, they  
are not strictly superior to the linear time operators for all problems.
Some problem characteristics may lead to superior performance with fewer
applications of one of the linear time mutation operators than if the constant time 
swap was instead used.

\begin{table}[H] 
\caption{Asymptotic runtime of cycle mutation and common permutation mutation operators.\label{tab:runtime}}
\begin{tabularx}{\textwidth}{CCC}
\toprule
\textbf{Mutation Operator}	& \textbf{Worst Case}	& \textbf{Average Case}\\
\midrule
$\mathrm{Cycle}(\mathit{kmax})$	& $O(\min(n,\mathit{kmax}^2))$	& $O(\min(n,\mathit{kmax}^2))$ \\
$\mathrm{Cycle}(\alpha)$	& $O(n)$	& $O(\min(n,(\frac{2 - \alpha}{1 - \alpha})^2))$ \\
$\mathrm{Swap}$	& $O(1)$	& $O(1)$ \\
$\mathrm{Insertion}$	& $O(n)$	& $O(n)$ \\
$\mathrm{Reversal}$	& $O(n)$	& $O(n)$ \\
$\mathrm{Scramble}$	& $O(n)$	& $O(n)$ \\
\bottomrule
\end{tabularx}
\end{table}

\subsection{New Measures of Permutation Distance}\label{sec:distance}

We now present three new measures of permutation distance: {\em cycle distance} 
(Section~\ref{sec:cycleDistance}), {\em cycle edit distance} (Section~\ref{sec:editDistance}),
and {\em $k$-cycle distance} (Section~\ref{sec:kdistance}).

\subsubsection{Cycle Distance}\label{sec:cycleDistance}

As a first step toward a distance measure appropriate for use in analyzing
fitness landscapes associated with the $\mathrm{Cycle}(\alpha)$ form of cycle mutation,
we present cycle distance. $\mathrm{Cycle}(\alpha)$ mutates a permutation
by inducing a cycle whose length is only limited by the permutation length itself.
Therefore, cycle distance is the count of the number of non-singleton
cycles. We can define the cycle distance between permutations $p_1$ and $p_2$ as:
\begin{equation}\label{eq:cycleDistance}
\delta(p_1, p_2) = \mathrm{CycleCount}(p_1, p_2) - \mathrm{FixedPointCount}(p_1, p_2) , 
\end{equation}
where $\mathrm{CycleCount}$ is the number of permutation
cycles, and $\mathrm{FixedPointCount}$ is the number of singleton
cycles or fixed points, which is a cycle of length one (i.e., a point where both permutations
contain the same element). We compute cycle distance in $O(n)$ time.

Cycle distance is a semi-metric, since it satisfies all of the metric properties except for 
the triangle inequality. First, it obviously satisfies non-negativity 
since $\delta(p_1, p_2)$ clearly cannot be negative since there cannot be
a negative number of non-singleton permutation cycles. Second, it is obvious 
that $\delta(p_1, p_2) = \delta(p_2, p_1)$, and is thus symmetric. Third, it satisfies 
the identity of indiscernibles as follows. When $p_1 = p_2 = p$, we have:
\begin{equation}
\delta(p, p) = \mathrm{CycleCount}(p, p) - \mathrm{FixedPointCount}(p, p) = n - n = 0 ,
\end{equation}
since the cycle count is $n$ (i.e., $n$ singleton cycles), and thus the number of fixed points is
also $n$. And in the other direction, we have:
\begin{equation}
\delta(p_1, p_2) = 0 \implies \mathrm{CycleCount}(p_1, p_2) = \mathrm{FixedPointCount}(p_1, p_2) \implies p_1 = p_2 ,
\end{equation}
since the only way that every cycle is a fixed point is if $p_1$ and $p_2$ are identical.

To demonstrate the violation of the triangle inequality, 
consider the permutations:
\begin{align}\label{eq:cycleNoTriangle}
\begin{split}
p_1 &= [0, 1, 2, 3, 4, 5, 6, 7, 8, 9] \\
p_2 &= [1, 0, 3, 2, 5, 4, 7, 6, 9, 8] \\
p_3 &= [0, 3, 2, 5, 4, 7, 6, 9, 8, 1] .
\end{split}
\end{align}
Observe $\delta(p_1, p_2) = 5$ since every consecutive pair of elements in
$p_1$ is swapped in $p_2$, creating five cycles of length two. 
All even numbered elements are fixed points for $p_1$ and $p_3$, and 
all odd numbered elements form a single cycle. That is, keep the even elements
fixed and cycle the odd elements to the left within $p_1$ to obtain $p_3$. Thus,
$\delta(p_1, p_3) = 1$. Finally, inspect $p_3$ and $p_2$ to note that if we cycle
all of the elements of $p_3$ one position to the right, we obtain $p_2$. Therefore,
$\delta(p_3, p_2) = 1$. Thus, since $\delta(p_1, p_2) > \delta(p_1, p_3) + \delta(p_3, p_2)$,
cycle distance does not satisfy the triangle inequality, and is only a semi-metric.

The diameter of the space of permutations $S_n$ of length $n$ given a measure of distance
$\delta$ is the maximal distance between points in that space.
Define the diameter $D(n, \delta)$ as:
\begin{equation}\label{eq:diameter}
D(n, \delta) = \max_{p_1,p_2 \in S_n} \{ \delta(p_1, p_2) \} .
\end{equation}

Since each non-singleton cycle contributes one to cycle distance, independent of cycle 
length, the maximum case is when the number of non-singleton cycles is maximized. The 
the smallest non-singleton cycle is length two. The maximum cycle distance 
therefore occurs when there 
are $\lfloor n/2 \rfloor$ cycles of length 2, leading to a diameter of:
\begin{equation}
D(n, \delta) = \left\lfloor \frac{n}{2} \right\rfloor .
\end{equation}

\subsubsection{Cycle Edit Distance}\label{sec:editDistance}

Although cycle distance relates to the $\mathrm{Cycle}(\alpha)$
operator, it is not actually an edit distance with $\mathrm{Cycle}(\alpha)$ as the edit operation.
However, we can utilize it to define such an edit distance. Define cycle edit 
distance as the minimum number of induced permutation cycles necessary to 
transform $p_1$ into $p_2$. We can formally define cycle edit distance as:
\begin{equation}\label{eq:editDistance}
\delta_{e}(p_1, p_2) = \begin{cases}
	0 	& \text{if $p_1 = p_2$} \\
	1	& \text{if $\delta(p_1, p_2) = 1$} \\
	2	& \text{if $\delta(p_1, p_2) > 1$} ,
	\end{cases}
\end{equation}
where $\delta(p_1, p_2)$ is cycle distance (Equation~\ref{eq:cycleDistance}).
If there is exactly one non-singleton cycle, we can trivially transform it to all fixed points
by cycling the elements of that cycle. If there are two or more non-singleton cycles,
there exists a cycle of the union of the elements of those cycles that will merge all of them
into a single larger cycle, possibly producing some fixed points. See Equation~\ref{eq:cycleNoTriangle} for an
example. Thus, one cycle edit operation merges all non-singleton cycles, and a second
cycle edit transforms it into all fixed points. We can compute cycle edit distance in $O(n)$ time since 
we can compute cycle distance in $O(n)$ time.

Cycle edit distance satisfies all of the metric properties as follows.
From Equation~\ref{eq:editDistance}, it is trivial to confirm non-negativity, symmetry, and
the identity of indiscernibles. Without loss of generality, assume that $p_1$, $p_2$, 
and $p_3$ are all different. Thus, 
$\delta_{e}(p_1, p_3) + \delta_{e}(p_3, p_2) \geq 1 + 1 \geq 2$, implying 
$\delta_{e}(p_1, p_2) \leq \delta_{e}(p_1, p_3) + \delta_{e}(p_3, p_2)$ since $\delta_{e}(p_1, p_2) \leq 2$
by definition. Thus, cycle edit distance satisfies the triangle inequality, and it is
therefore a metric.

Multiple non-singleton cycles can only exist if permutation length $n \geq 4$,
and permutations of length one must be identical. Thus, 
the diameter of cycle edit distance is:
\begin{equation}
D(n, \delta_{e}) = \begin{cases}
	0 & \text{if $n \leq 1$} \\
	1 & \text{if $1 < n \leq 3$} \\
	2 & \text{if $3 < n$} .
	\end{cases}
\end{equation}

\subsubsection{$K$-Cycle Distance}\label{sec:kdistance}

Cycle distance and cycle edit distance assume that arbitrary length 
cycles can be induced in a single operation, which is needed
for $\mathrm{Cycle}(\alpha)$ since it does not restrict
the induced cycle length. However, this is not suitable when
characterizing fitness landscapes for
the $\mathrm{Cycle}(\mathit{kmax})$ mutation operator, which does
limit the cycle length to $\mathit{kmax}$.

We now define $k$-cycle distance, which 
limits operations to inducing cycles of 
lengths $\{2, 3, \ldots, k\}$. The $k$-cycle distance is not an edit distance.
Instead, it is a weighted sum over the cycles of the
pair of permutations, where the weight of each cycle is the minimum number
of induced cycles of length at most $k$ necessary to transform the cycle to all
fixed points. That is, $k$-cycle distance does not consider operations that 
span multiple cycles.   

A cycle of length $c \leq k$ can be transformed
to $c$ fixed points by inducing a cycle of length $c$. If 
the cycle length $c > k$, a sequence of cycles can derive
$c$ fixed points. For $k$-cycle distance,
iteratively induce cycles of length $k$ until the remaining cycle
length is at most $k$, completing the transformation with one final cycle operation.
Each intermediate cycle creates $k-1$ fixed points. Consider the following example beginning with
permutations:
\begin{align}
\begin{split}
p_1 &= [0, 1, 2, 3, 4, 5, 6, 7, 8, 9] \\
p_2 &= [2, 3, 1, 5, 6, 7, 8, 9, 4, 0] .
\end{split}
\end{align}
There are two non-singleton permutation cycles in this example
with element sets:
\begin{equation}
\{ 1, 3, 5, 7, 9, 0, 2 \}, \{ 4, 6, 8 \} .
\end{equation}
If we are computing $3$-cycle distance, then we consider
cycle mutations of lengths 2 and 3. The cycle with elements $\{ 4, 6, 8 \}$
can thus be transformed into all fixed points with a single cycle mutation
since its length is less than or equal to 3 to obtain:
\begin{align}
\begin{split}
p_1 &= [0, 1, 2, 3, 4, 5, 6, 7, 8, 9] \\
p_2 &= [2, 3, 1, 5, 4, 7, 6, 9, 8, 0] .
\end{split}
\end{align}
The longer cycle $\{ 1, 3, 5, 7, 9, 0, 2 \}$ requires
a sequence of three operations. The first two produce $k-1=2$ fixed 
points each with one final cycle mutation for the 
remaining three elements. First, cycle
elements 3, 5, and 7, resulting in fixed points 
for elements 3 and 5:
\begin{align}
\begin{split}
p_1 &= [0, 1, 2, 3, 4, 5, 6, 7, 8, 9] \\
p_2 &= [2, 7, 1, 3, 4, 5, 6, 9, 8, 0] .
\end{split}
\end{align}
A cycle of $\{ 1, 7, 9, 0, 2 \}$
remains. Cycle elements 7, 9, and 0, creating fixed points
7 and 9:
\begin{align}
\begin{split}
p_1 &= [0, 1, 2, 3, 4, 5, 6, 7, 8, 9] \\
p_2 &= [2, 0, 1, 3, 4, 5, 6, 7, 8, 9] .
\end{split}
\end{align}
A cycle of $\{ 1, 0, 2 \}$ remains, which transforms into
fixed points with one final cycle mutation:
\begin{align}
\begin{split}
p_1 &= [0, 1, 2, 3, 4, 5, 6, 7, 8, 9] \\
p_2 &= [0, 1, 2, 3, 4, 5, 6, 7, 8, 9] .
\end{split}
\end{align}

With this example in mind, we now derive the $k$-cycle distance, assuming
$k \geq 2$. First, given a cycle of length $c$, the fewest cycle edits of length at most
$k$ necessary to transform it to $c$ fixed points is $\lceil (c - 1)/(k - 1) \rceil$.
To compute $k$-cycle distance, sum this over
all cycles, $\mathrm{CycleSet}(p_1,p_2)$, of the permutations $p_1$ and $p_2$. Therefore, define $k$-cycle distance by:
\begin{equation}\label{eq:KCycleDistance}
\delta_{k}(p_1,p_2) = \sum_{\substack{\mathrm{cycle} \in \mathrm{CycleSet}(p_1,p_2) \\ c = \left|\mathrm{cycle}\right|}} \left\lceil \frac{c - 1}{k - 1} \right\rceil .
\end{equation} 

The $k$-cycle distance can be computed in $O(n)$ time.
It is a metric when $k \leq 4$, but it is only a semi-metric for
$k \geq 5$. Independent of $k$, it trivially 
satisfies non-negativity since the expression within the sum of Equation~\ref{eq:KCycleDistance} 
is never negative; and since $\mathrm{CycleSet}(p_1,p_2)$ computes the same set of cycles 
as $\mathrm{CycleSet}(p_2,p_1)$, it is also trivially symmetric. It satisfies 
the identity of indiscernibles as follows. If $p_1 = p_2 = p$, then there 
are $n$ singleton cycles, and since the expression within the sum of 
Equation~\ref{eq:KCycleDistance} is 0 when $c=1$, we have $\delta_{k}(p,p) = 0$. In the 
other direction, if $\delta_{k}(p_1,p_2) = 0$, 
then all elements in the summation must be 0, and that only occurs with fixed points. 
Thus, $\delta_{k}(p_1,p_2) = 0 \implies p_1 = p_2$. 

The triangle inequality is violated if $k \geq 5$. Consider this case 
with $k = 6$:
\begin{align}
\begin{split}
p_1 &= [0, 1, 2, 3, 4, 5] \\
p_2 &= [1, 0, 3, 2, 5, 4] \\
p_3 &= [0, 3, 2, 5, 4, 1] .
\end{split}
\end{align}
The $6$-cycle distance allows cycle operations of length up to six.
Note that $\delta_{6}(p_1,p_2) = 3$ since there are three 
cycles of length two; $\delta_{6}(p_1,p_3) = 1$ since
the even numbered elements are fixed points, and the odd numbered elements
form a single cycle of length three; and $\delta_{6}(p_3,p_2) = 1$, with a single
cycle of length six (i.e., cycle the elements of $p_3$ one position to the right
to obtain $p_2$). Thus, 
$\delta_{6}(p_1,p_3) + \delta_{6}(p_3,p_2) = 1 + 1 = 2 < \delta_{6}(p_1,p_2)$.
Therefore, $6$-cycle distance is only a semi-metric, as is $k$-cycle distance
for any other $k \geq 6$.

We can produce a similar example for the case of $k=5$ as follows:
\begin{align}
\begin{split}
p_1 &= [0, 1, 2, 3, 4, 5] \\
p_2 &= [1, 0, 3, 2, 5, 4] \\
p_3 &= [0, 3, 2, 5, 1, 4] .
\end{split}
\end{align}
As in the previous example, $\delta_{5}(p_1,p_2) = 3$. Note that
$\delta_{5}(p_1,p_3) = 1$ since $\{0, 2\}$ are fixed points, and there
is a single cycle of the elements $\{1, 3, 5, 4\}$;
and $\delta_{5}(p_3,p_2) = 1$, with a fixed point for
element 4, and a single cycle of the remaining five elements. 
In this example, 
$\delta_{5}(p_1,p_3) + \delta_{5}(p_3,p_2) < \delta_{5}(p_1,p_2)$. Thus,
$5$-cycle distance is also only a semi-metric.

In general, as many as $k$ non-singleton cycles can be merged into a
single larger cycle by cycling $k$ elements,
where the cycle edit includes at least one element of each of the merged 
cycles. This is also true when $k < 5$, but for $k < 5$ the 
resulting merged cycle is larger than $k$, requiring multiple cycle edits
to transform into all fixed points. 

When $k \leq 4$, the $k$-cycle distance satisfies the triangle inequality,
and is a metric; but when $k \geq 5$, it is only a semi-metric.
Indeed, when $k \leq 4$, the $k$-cycle distance is equivalent to an edit
distance with cycles of length up to $k$ as the edit operations. For example, 
$2$-cycle distance is equivalent to an existing distance metric
known as interchange distance, which is an edit distance with swap as its 
edit operation, which is a metric. 

The diameter of the space of permutations for $k$-cycle distance
depends upon the permutation length $n$ in relation to
$k$. When $k$ is high, distance is maximized by maximizing the number
of permutation cycles, which occurs when there are $\lfloor n/2 \rfloor$  
cycles of length 2. When $k$ is low, distance is maximized when there is
a single cycle of length $n$, which requires 
$\lceil (n - 1)/(k - 1) \rceil$ cycle operations to transform to $n$ fixed points.
Thus, the diameter is:
\begin{equation}
D(n, \delta_{k}) = \max \left\{ \left\lfloor \frac{n\vphantom{-0}}{2\vphantom{-0}} \right\rfloor, \, \left\lceil \frac{n-1}{k-1} \right\rceil \right\} .
\end{equation}

\subsection{Fitness Landscape Analysis}\label{sec:landscapes}

The fitness landscape analysis includes calculation of landscape diameters
in Section~\ref{sec:diameter}, FDC analysis in
Section~\ref{sec:fdc}, and an analysis of the search landscape calculus 
in Section~\ref{sec:calculus}. We synthesize the fitness landscape analysis 
findings in Section~\ref{sec:findings}.

When an edit distance is required, such as to compute
FDC and the diameter, we use cycle edit distance for $\mathrm{Cycle}(\alpha)$, and 
$k$-cycle distance for $\mathrm{Cycle}(\mathit{kmax})$.
Swap's edit distance is interchange distance, the 
minimum number of swaps to transform $p_1$ into $p_2$:
\begin{equation}\label{eq:xchange}
\delta_{i}(p_1,p_2) = n - \mathrm{CycleCount}(p_1,p_2) ,
\end{equation}
where $\mathrm{CycleCount}(p_1,p_2)$ is the number of permutation cycles. The edit distance 
for insertion mutation is known as reinsertion distance, the
minimum number of insertion mutations needed to transform $p_1$ into $p_2$.
It is efficiently computed as~\cite{Cicirello2016}: 
\begin{equation}\label{eq:reinsert}
\delta_{r}(p_1,p_2) = n - \lvert\mathrm{LongestCommonSubsequence}(p_1,p_2)\rvert ,
\end{equation}
where $\mathrm{LongestCommonSubsequence}(p_1,p_2)$ is the longest common 
subsequence.
It is not feasible to utilize an edit distance to analyze reversal mutation
landscapes, because computing reversal edit distance is NP-Hard~\cite{caprara1997}.
Instead we utilize cyclic edge distance~\cite{ronald1997}, which interprets a permutation as a 
cyclic sequence of edges (e.g., 1 following 2 is treated as an edge between 1 and 2)
and counts the edges in $p_1$ that are not in $p_2$. For scramble mutation,
a trivial edit distance from a permutation to itself is 0, and 
to any other permutation is 1, since any permutation is reachable by some
shuffling of any other.

\subsubsection{Fitness Landscape Diameter}\label{sec:diameter}

Table~\ref{tab:diameter} compares the diameters of fitness landscapes for
the various mutation operators. Except where noted, we use the metrics 
identified above. The diameter of both swap and insertion landscapes is $n - 1$. 
The maximum distance for swap occurs for a single $n$ element cycle. 
The maximum case for insertion occurs when one permutation is the reverse of the 
other. For reversal mutation, we use the exact diameter
for a reversal edit distance rather than relying on the surrogate distance measure
identified earlier. The diameter of a reversal landscape is 
$n - 1$, proven by Bafna and Pevzner~\cite{Bafna1996}, the proof 
of which is well beyond the scope of this article. The diameter of a scramble 
landscape is simply 1 since every permutation is reachable from any other via a single 
scramble operation. 

\begin{table}[H] 
\caption{Fitness landscape diameter for cycle mutation and common permutation mutation operators.\label{tab:diameter}}
\begin{tabularx}{\textwidth}{CC}
\toprule
\textbf{Mutation Operator}	& \textbf{Diameter}	\\
\midrule
$\mathrm{Cycle}(\mathit{kmax}),\,\mathit{kmax}\geq5$	& $\approx 2n/\mathit{kmax}$ \\
$\mathrm{Cycle}(\mathit{kmax}),\,\mathit{kmax}\leq4$	& $\max \left\{ \left\lfloor n/2 \right\rfloor, \, \left\lceil (n - 1)/(\mathit{kmax} - 1) \right\rceil \right\}$	 \\
$\mathrm{Cycle}(\alpha)$	& $2$	 \\
$\mathrm{Swap}$	& $n - 1$ \\
$\mathrm{Insertion}$	& $n - 1$ \\
$\mathrm{Reversal}$	& $n - 1$ \\
$\mathrm{Scramble}$	& $1$	 \\
\bottomrule
\end{tabularx}
\end{table}

The diameter of a $\mathrm{Cycle}(\alpha)$ landscape is
2, which is the diameter of the space of permutations for cycle edit distance 
(Section~\ref{sec:editDistance}). When $\mathit{kmax} \leq 4$, the 
diameter of $\mathrm{Cycle}(\mathit{kmax})$ is 
$\max \left\{ \left\lfloor n/2 \right\rfloor, \, \left\lceil (n - 1)/(\mathit{kmax} - 1) \right\rceil \right\}$,
which is the maximum $k$-cycle distance (Section~\ref{sec:kdistance}).

We previously saw that $k$-cycle distance does not satisfy the triangle inequality
if $\mathit{kmax} \geq 5$. Although computing a $\mathrm{Cycle}(\mathit{kmax})$ edit distance
for $\mathit{kmax} \geq 5$ is too costly for our purpose, it is straightforward to
compute its diameter. Recall the examples illustrating that $k$-cycle distance
violates the triangle inequality, and specifically that two cycle edits of length $k$
can transform a set of cycles with a total of $k$ elements into $k$ fixed points. Thus,
the maximum case of $\lfloor n/2 \rfloor$ cycles of length 2 can be transformed to $n$
fixed points with approximately $2n/\mathit{kmax}$ applications of the $\mathrm{Cycle}(\mathit{kmax})$
operator when $\mathit{kmax} \geq 5$.

\subsubsection{Fitness Distance Correlation}\label{sec:fdc}

We now compute FDC for small instances of the TSP, LCS, and QAP. 
To compute FDC for a problem instance, you need all optimal solutions 
to that instance. This necessitates utilizing an instance small enough to 
feasibly and reliably determine all optimal
solutions. The FDC calculations in this section use a permutation length $n=10$ for this reason,
which means a 10-city TSP, an LCS with graphs of 10 vertexes, and a QAP with 10 by 10
cost and distance matrices. The EA experiments later in the paper use larger problem
instances to experimentally compare the performance of the different mutation operators. 

We use a TSP instance with 10 cities arranged equidistantly
around a circle of radius 10, and Euclidean distance for the edge costs. In this way,
the optimal solutions are known a priori. There are 20 optimal permutations,
all representing the same TSP tour following the cities around the
circle, including ten possible starting cities and two travel directions.

We generate a QAP instance such that a random permutation $p$ of length $10$
is the known optimal solution. Let $A = [ (1, 90), (2, 89), \ldots, (90, 1)]$. 
Shuffle this array of ordered pairs. The $90$
non-diagonal elements of the 10 by 10 cost matrix $C$ are populated row by row using the 
first element of the tuples in $A$. Let $(u,v)$ be one of these tuples. If $C[i][j]$
is set to $u$, then $D[p(i)][p(j)]$ in the distance matrix $D$ is set to $v$.

\begin{figure}[H]
\begin{tikzpicture} 
    \begin{scope} [vertex style/.style={draw, circle, minimum size=4mm, inner sep=0pt, outer sep=0pt, shade, top color=blue, bottom color=purple}] 
      \path \foreach \i in {0,...,4}{%
       (72*\i:0.75) coordinate[vertex style] (a\i)
       (72*\i:1.5) coordinate[vertex style] (b\i)}
       ; 
    \end{scope}
     \begin{scope} [edge style/.style={draw=red,double=brown}]
       \foreach \i in {0,...,4}{%
       \pgfmathtruncatemacro{\nextb}{mod(\i+1,5)}
       \pgfmathtruncatemacro{\nexta}{mod(\i+2,5)} 
       \draw[edge style] (a\i)--(b\i);
       \draw[edge style] (a\i)--(a\nexta);
       \draw[edge style] (b\i)--(b\nextb);
       }  
     \end{scope}
  \end{tikzpicture}
\caption{The Petersen graph, a strongly regular graph.\label{fig:petersen}}
\end{figure}
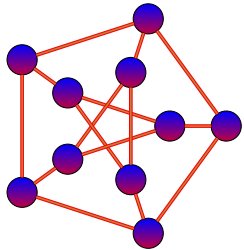

We generate an LCS instance consisting of a pair of isomorphic graphs.
In this way, we know that the LCS is either of those graphs or any
of the other automorphisms of the graph. We use a strongly regular graph in the analysis.
Strongly regular graphs are especially challenging for algorithms for the LCS and other
related problems. Specifically, we use the Petersen graph~\cite{Harary1967}, 
shown in Figure~\ref{fig:petersen}, which has 120 automorphisms,
and thus 120 optimal vertex mappings. Such instances are hard because many 
vertexes of a strongly regular graph look the same locally to a solver, which can be simultaneously
attracted toward different distinct solutions in a space plagued by complex local optima.

Although guidance on interpreting FDC varies, an 
$r \leq -0.15$ is commonly considered an easier fitness landscape 
since it implies that fitness increases as distance to an optimal solution 
decreases, while $r \geq 0.15$ is likely a deceptive landscape since 
fitness increases as you move away from optimal solutions, and 
$-0.15 < r < 0.15$ is considered difficult since there is very little 
correlation between fitness and distance to an optimal 
solution~\cite{fdc}.

For each combination of problem and mutation operator, we compute FDC exactly, 
calculating Pearson correlation over all $n! = 3628800$ permutations of 
length $n=10$. Table~\ref{tab:fdc} summarizes the results. Without loss of generality,
for TSP and QAP we compute FDC from the cost function (to minimize) 
rather than a fitness function (to maximize), flipping the
sign of the FDC with positive FDC implying
more straightforward problem solving. Since LCS is a maximization problem, the sign
of the FDC is interpreted normally.

\begin{table}[H] 
\caption{FDC for combinations of mutation operator and problem.\label{tab:fdc}}
\begin{tabularx}{\textwidth}{CCCC}
\toprule
\textbf{Mutation Operator}	& \textbf{TSP} & \textbf{QAP} & \textbf{LCS}	\\
\midrule
$\mathrm{Cycle}(\alpha)$ & -0.0569	& 0.0213	& -0.0278 \\
$\mathrm{Cycle}(5)$		& 0.1801	& 0.1339	& -0.5342 \\
$\mathrm{Cycle}(4)$		& 0.1667	& 0.1737	& -0.3984 \\
$\mathrm{Cycle}(3)$		& 0.2482	& 0.2210	& -0.6180 \\
$\mathrm{Swap}$			& 0.3318	& 0.2245	& -0.6355 \\
$\mathrm{Insertion}$	& 0.5277	& 0.0305	& -0.3547 \\
$\mathrm{Reversal}$		& 0.8459	& 0.0189	& -0.0350 \\
$\mathrm{Scramble}$		& 0.0117	& 0.0048	& -0.0340 \\
\bottomrule
\end{tabularx}
\end{table}

For the TSP, there is very strong FDC for reversal mutation, and lesser
but still strong FDC for insertion mutation. The FDC analysis predicts 
that these will perform better than the others. Swap 
has the next highest FDC. Although not as 
high, all cases of $\mathrm{Cycle}(\mathit{kmax})$ exhibit $r > 0.15$. 
Given the correlation strength of reversal and insertion landscapes, we 
expect them to dominate the others, but swap and
$\mathrm{Cycle}(\mathit{kmax})$ may also perform well.

The strength of the correlations are not nearly as high for the QAP,
and only three of the mutation operators have an $r \geq 0.15$, including
swap mutation and $\mathrm{Cycle}(\mathit{kmax})$ with a $\mathit{kmax}$
of either 3 or 4. This may be a problem where cycle mutation is better suited. 

Because LCS is a maximization problem, FDC should be interpreted in the ordinary sense
with FDC computed from fitness where higher fitness implies better solutions. Thus, negative FDC
implies easier problem solving. In this case, there
is very strong FDC for swap mutation, $\mathrm{Cycle}(3)$ mutation, and
$\mathrm{Cycle}(5)$ mutation. Both $\mathrm{Cycle}(4)$ and 
insertion mutation also have $r \leq -0.15$, so should also progressively
lead toward the solution.

The FDC for scramble mutation is very near 0 for all three problems.
Scramble is thus unlikely to perform well. The FDC 
for $\mathrm{Cycle}(\alpha)$ is also very near 0 for all three 
problems. However, we will see that FDC misses
an important behavioral property of this operator.

\subsubsection{Search Landscape Calculus}\label{sec:calculus}

Let $\eta(p)$ be the neighbors of $p$ (i.e., solutions 
reachable with one mutation) and $f(p)$ is fitness. Search landscape 
calculus~\cite{Cicirello2016} defines the average 
local rate of fitness change:
\begin{equation}
\Delta[f](p) = \frac{1}{\left| \eta(p) \right|} \sum_{p^{\prime} \in \eta(p)} \left| f(p) - f(p^{\prime}) \right| .
\end{equation}

Search landscape calculus then defines $\Delta[f]$ as the average of $\Delta[f](p)$ over all $p$.
It is infeasible to directly compute $\Delta[f]$ for the true fitness 
function $f$. Therefore, the search landscape calculus focuses on 
topological properties that influence fitness, such as absolute positions, 
relative positions, and element precedences for permutation problems, replacing  
$f$ by a distance function $\delta$ relevant to the context of 
a problem feature. Thus, define $\Delta[\delta](p)$:
\begin{equation}
\Delta[\delta](p) = \frac{1}{\left| \eta(p) \right|} \sum_{p^{\prime} \in \eta(p)} \delta(p, p^{\prime})  ,
\end{equation}
and from this derive $\Delta[\delta]$ as the average of $\Delta[\delta](p)$ over all points $p$.
The distance $\delta$ must correspond to the topological property under analysis, 
and its maximum must be proportional to the permutation length $n$ 
(i.e., $\max \{ \delta(p_1, p_2) \} \in \Theta(n)$). 

We focus on two topological features, absolute element positions and
edges, and thus require corresponding distance functions. 
We use exact match distance~\cite{ronald1998} (denoted as $\delta_{em}$), 
which is the count of the number of positions containing different elements,
as a measure of how different two permutations are within the context of
absolute positioning; and we use cyclic edge distance~\cite{ronald1997}
(denoted as $\delta_{ce}$) as a measure of how different two permutations 
are within the context of relative positions. Both meet the requirement
that the maximum distance is proportional to $n$. It is exactly 
$n$ in both cases.

Table~\ref{tab:calculus} summarizes $\Delta[\delta_{em}]$ and $\Delta[\delta_{ce}]$
for both forms of cycle mutation, and the other mutation operators considered.
The rates of change of the topological properties of
swap, insertion, reversal, and scramble were determined elsewhere~\cite{Cicirello2016}.
The $\Delta[\delta_{em}]$ is the average number of elements whose
absolute positions are changed by a single application of the operator.
For cycle mutation this is the average length of the induced cycle, 
determined earlier in the article---$\mathrm{Cycle}(\alpha)$ in 
Section~\ref{sec:alpha} and $\mathrm{Cycle}(\mathit{kmax})$ in Section~\ref{sec:kmax}.
To compute $\Delta[\delta_{ce}]$ we need the average number of edges changed by
the operator. For cycle mutation, this depends upon the cycle length, and it also depends upon
whether any cycle elements are adjacent. As the permutation length $n$ increases, the probability of
adjacent cycle elements decreases. For sufficiently large $n$, the probability of adjacent cycle
elements approaches 0, in which case the number of edges replaced by cycle mutation 
is on average twice the length of the cycle. Thus, $\Delta[\delta_{ce}] = 2 \Delta[\delta_{em}]$
for both forms of cycle mutation.

\begin{table}[H] 
\caption{Average rates of change of fitness landscape topological properties.\label{tab:calculus}}
\begin{tabularx}{\textwidth}{CCC}
\toprule
\textbf{Mutation Operator}	& \textbf{$\Delta[\delta_{em}]$} & \textbf{$\Delta[\delta_{ce}]$}	\\
\midrule
$\mathrm{Cycle}(\alpha)$ & $(2 - \alpha)/(1 - \alpha)$	& $(4 - 2\alpha)/(1 - \alpha)$	  \\
$\mathrm{Cycle}(\mathit{kmax})$		& $(\mathit{kmax}+2)/2$	& $\mathit{kmax}+2$ 	  \\
$\mathrm{Swap}$			& $2$	&  $4$   \\
$\mathrm{Insertion}$	& $(n+4)/3$	& $3$	 \\
$\mathrm{Reversal}$		& $[(n+1)/3,(n+4)/3]$	& $2$	 \\
$\mathrm{Scramble}$		& $(n+1)/3$	& $(n+1)/3$ 	 \\
\bottomrule
\end{tabularx}
\end{table}

In the search landscape calculus, when $\lim_{n \to \infty} \Delta[\delta] = \infty$
the fitness landscape exhibits very large local changes in fitness, often due 
to many deep local optima that are difficult to escape. Thus, we expect poor performance
from insertion, reversal, and scramble on absolute-positioning problems
since $\Delta[\delta_{em}]$ grows with $n$, as is the case for scramble
when relative positions are more important.
When $\lim_{n \to \infty} \Delta[\delta] = C$, for a non-zero finite
constant $C$, the search landscape calculus suggests that the landscape is
smooth locally. Due to constant $\Delta[\delta_{em}]$, swap and 
cycle mutation should perform better than the others for problems like LCS and
QAP where absolute positions more greatly impact fitness. 
All operators except for scramble have constant $\Delta[\delta_{ce}]$, and 
thus potentially relevant to problems like the TSP where edges influence fitness
more than absolute element positions.

For cycle mutation, it should be noted that its topological characteristics
depend upon the specific parameter settings. For example, higher values 
of $\mathit{kmax}$ likely lead to the same sort of disruption inherent
in scramble mutation, as would values of $\alpha$ very near 1.0.

Previously, we saw that FDC suggested that $\mathrm{Cycle}(\alpha)$
would likely perform poorly for all three problems due to FDC very near 0; 
whereas the search landscape calculus suggests that it may be relevant for all 
three problems. We now reconcile the discrepancy between these two fitness landscape 
analysis tools. The number of permutations within one step
of a given permutation with respect to $\mathrm{Cycle}(\alpha)$  
is enormous, leading to extremely low variation in distance and then subsequently to near 0
FDC. However, a large proportion of the $\mathrm{Cycle}(\alpha)$ neighbors
are very low probability events. Thus, most of the time it behaves more
like $\mathrm{Cycle}(\mathit{kmax})$ with a very low $\mathit{kmax}$, but with the ability
to make larger jumps.

\subsubsection{Summary of Fitness Landscape Analysis Findings}\label{sec:findings}

Due to strong FDC, we expect reversal mutation to dominate the others for the TSP, 
finding lower cost solutions with the same or fewer evaluations. Insertion should
likely be the next best since it also has strong FDC, and a low constant rate 
of fitness change for edge-focused problems. Swap and cycle mutation, if configured to 
emphasize smaller cycles, may also be relevant for such problems. Due to strong FDC 
for QAP and LCS, and low constant rates of change of exact match 
distance, we anticipate swap and both forms of cycle mutation will find lower cost
solutions for absolute-positioning focused problems.

\section{Results}\label{sec:experiments}

We experimentally compare the two forms of cycle 
mutation with commonly encountered permutation mutation operators. 
Our experiments are implemented in Java. Our test machine is a 
Windows 10 PC, with an AMD A10-5700 3.4 GHz CPU, and 8 GB memory. The 
code is compiled with OpenJDK 17.0.2 for a Java 11 target, and runs 
on an OpenJDK 64-bit Server VM Temurin-17.0.2+8. 
The code is open source, licensed via the GPL 3.0, and available 
on GitHub at \url{https://github.com/cicirello/cycle-mutation-experiments},
and includes the code to analyze the experiment data and to generate 
the figures.

In the experiments, we apply each of the mutation operators within 
both a $(1+1)$-EA as well as in an SA. In 
a $(\mu+\lambda)$-EA~\cite{Eiben2015}, the population size is $\mu$, 
$\lambda$ offspring are created in each generation, and the 
best $\mu$ individuals from the combination of parents and
offspring survive into the next generation. The $(1+1)$-EA
is commonly used in experimental studies. We employ it 
here since it removes the impact of choice of selection operator
and crossover operator, and also eliminates many parameters such
as population size, mutation rate, etc. Additionally, it supports a 
more direct comparison with the non-population approach of SA. For
the SA, we use the parameter-free Self-Tuning Lam
Annealing~\cite{cicirello2021} that adaptively adjusts the temperature parameter
based on problem-solving feedback. 

We consider three cases of $\mathrm{Cycle}(\mathit{kmax})$ mutation, including
$\mathrm{Cycle}(3)$, $\mathrm{Cycle}(4)$, and $\mathrm{Cycle}(5)$; and
three cases of $\mathrm{Cycle}(\alpha)$ mutation, including
$\mathrm{Cycle}(0.25)$, $\mathrm{Cycle}(0.5)$, and $\mathrm{Cycle}(0.75)$.
All results are 50 run averages. 
We use run lengths $\{ 10^{2}, 10^{3}, 10^{4}, 10^{5}, 10^{6}, 10^{7} \}$ 
in number of evaluations. We test significance with the Wilcoxon rank sum test.
The results on the TSP, QAP, and LCS are presented in 
sections~\ref{sec:tsp},~\ref{sec:qap}, and~\ref{sec:lcs}, respectively.

\begin{figure}[H]
\begin{adjustwidth}{-\extralength}{0cm}
\begin{tabular}{cc}
\includegraphics[width=3.47 in]{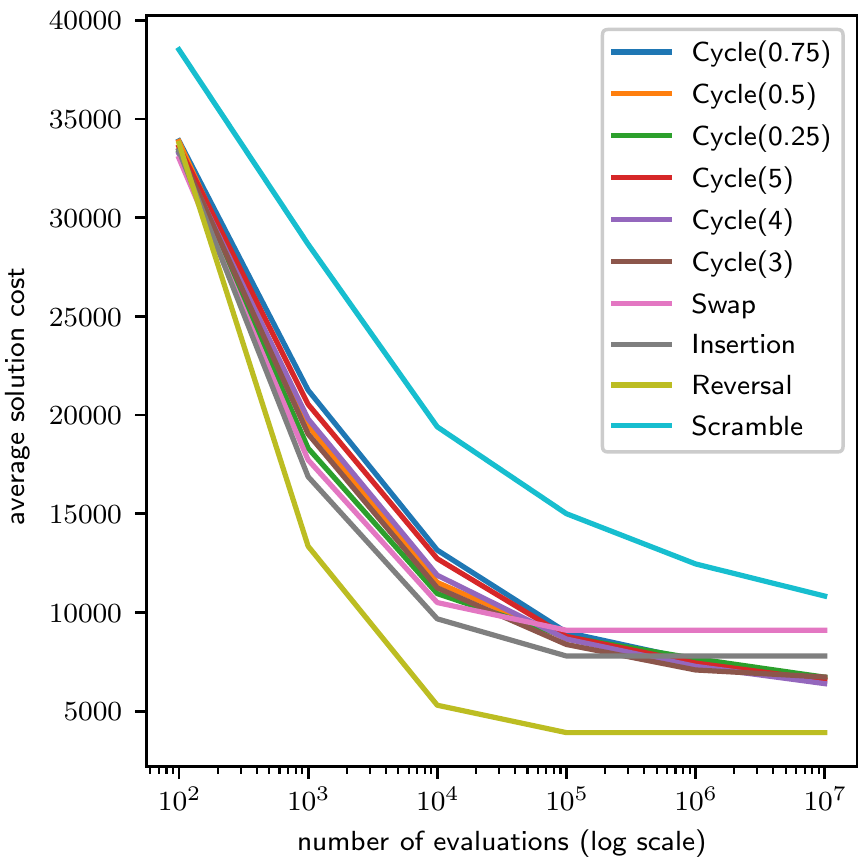} &
\includegraphics[width=3.47 in]{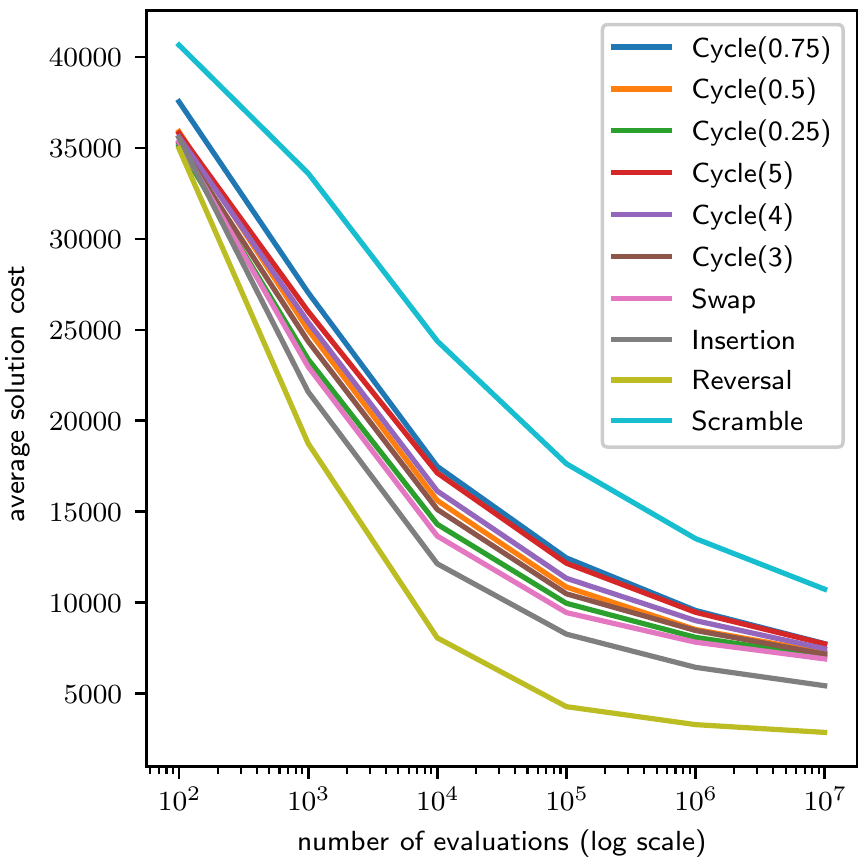} \\
(\textbf{a}) & (\textbf{b})
\end{tabular}
\end{adjustwidth}
\caption{Results on the TSP for: (\textbf{a}) $(1+1)$-EA, and (\textbf{b}) SA.\label{fig:tsp}}
\end{figure}

\subsection{TSP Results}\label{sec:tsp}

Each of the 50 TSP instances consists of 100 cities. An instance
is defined by a random distance matrix, such that the distance between cities
is a uniformly random integer from the interval $[1, 1000]$. The results are shown
in Figure~\ref{fig:tsp} with number of evaluations on the
horizontal axis at log scale, and average solution cost over 50 runs on the
vertical axis.

Consistent with the extremely strong FDC that we earlier observed for reversal
mutation for the TSP, we find that reversal is clearly dominant on the 100-city 
TSP instances within both the EA (Figure~\ref{fig:tsp}(a)) and 
SA (Figure~\ref{fig:tsp}(b)), finding lower cost solutions with fewer evaluations. 
All comparisons to reversal mutation are extremely statistically 
significant (e.g., Wilcoxon rank sum test p-values very near 0) except for the 
shortest 100 evaluation runs. For SA, the second best mutation operator is 
insertion (results also statistically significant). We earlier saw that insertion 
had the second strongest FDC for the TSP. Swap and the various cases of cycle 
mutation all find solutions of approximately equivalent cost within the SA, especially 
for very long run lengths.

The EA comparison (Figure~\ref{fig:tsp}(a)) is a bit more interesting. Although 
for mid-length runs, insertion is second best at statistically significant levels, 
the various cases of cycle mutation surpass insertion mutation for the longest runs. 
Insertion mutation exhibits a premature convergence effect, converging to a significantly
suboptimal solution $10^5$ evaluations into the EA runs, as does swap. However,  
cycle mutation continues to find lower cost solutions, overtaking insertion mutation.
The convergence effect in the EA is due to the smaller neighborhoods of swap 
and insertion, leading to greater impact of local optima. Cycle mutation has a larger 
neighborhood so better avoids this, continuing to show progress. In fact, even though 
we saw near-zero FDC for $\mathrm{Cycle}(\alpha)$, all cases of $\mathrm{Cycle}(\alpha)$ 
continue to make progress, although converging at a slower rate than reversal.

\subsection{QAP Results}\label{sec:qap}

Each of the 50 QAP instances consists of a 50 by 50 cost matrix and a 
50 by 50 distance matrix, with integer costs and distances generated 
uniformly at random from $[1, 50]$.

\begin{figure}[H]
\begin{adjustwidth}{-\extralength}{0cm}
\begin{tabular}{cc}
\includegraphics[width=3.47 in]{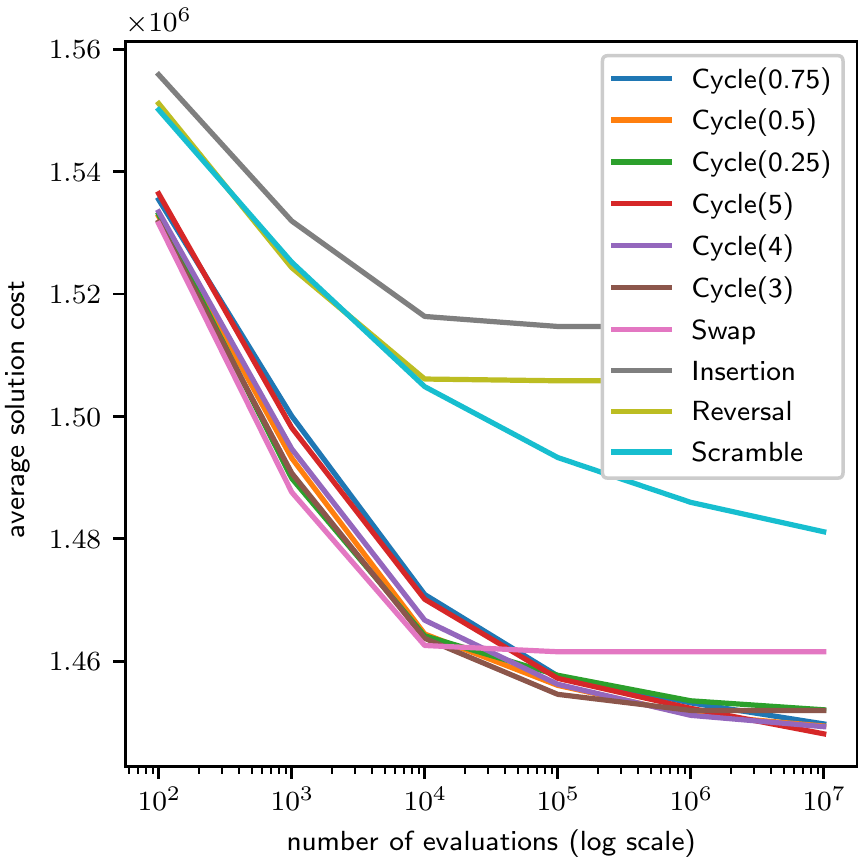} &
\includegraphics[width=3.47 in]{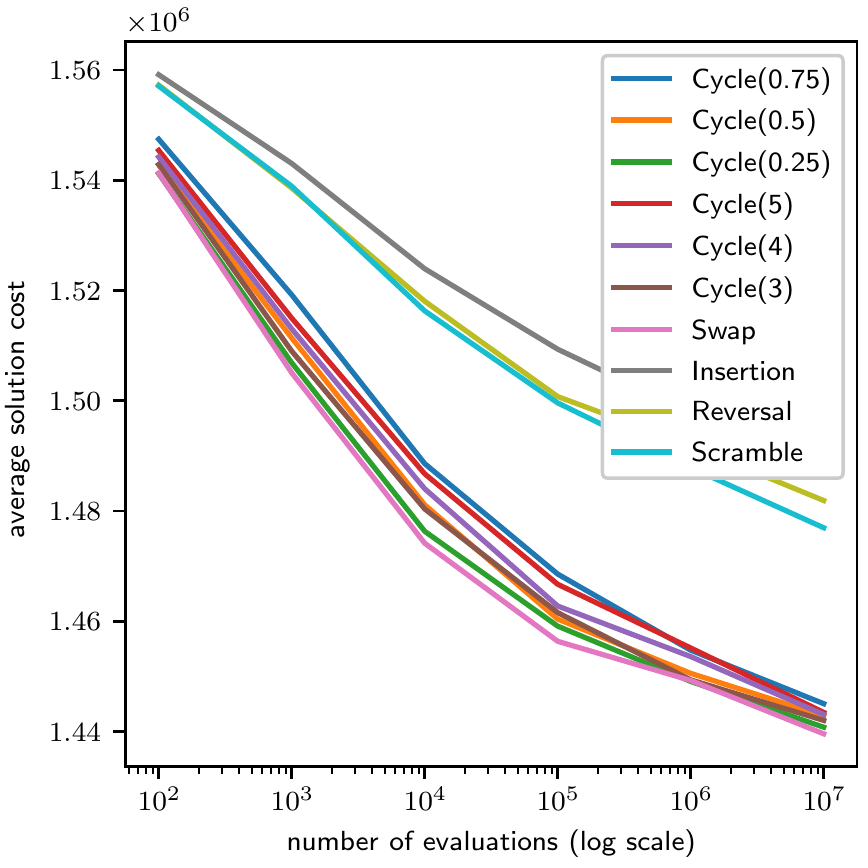} \\
(\textbf{a}) & (\textbf{b})
\end{tabular}
\end{adjustwidth}
\caption{Results on the QAP for: (\textbf{a}) $(1+1)$-EA, and (\textbf{b}) SA.\label{fig:qap}}
\end{figure}

The results are visualized in Figure~\ref{fig:qap}. We earlier saw that 
FDC suggested that swap, $\mathrm{Cycle}(3)$, 
and $\mathrm{Cycle}(4)$ would likely perform better than the others, 
while the search landscape calculus suggested that swap and all forms
of cycle mutation are appropriate for the QAP. Consistent with the fitness 
landscape analysis, scramble, reversal, and insertion all perform poorly 
for QAP within both the EA and SA. For the longest SA runs (Figure~\ref{fig:qap}(b)), 
there is little difference in solution cost among swap and the various cases of cycle 
mutation. For runs of $10^4$ to $10^5$ SA evaluations, swap and 
$\mathrm{Cycle}(0.25)$ find solutions with lower values of the cost function
than the others at statistically significant levels.

The EA results are especially interesting (Figure~\ref{fig:qap}(a)). Swap 
suffers from a premature convergence effect at $10^4$ evaluations, while
all cycle mutation cases continue to make substantial progress minimizing the
cost function. Furthermore, $\mathrm{Cycle}(0.25)$ and $\mathrm{Cycle}(3)$'s rates
of convergence are equivalent to swap up to that point. The superior convergence
effect is attributed to cycle mutation's larger neighborhood, especially in the case 
of $\mathrm{Cycle}(\alpha)$. We don't see the same behavior with SA (Figure~\ref{fig:qap}(b)) 
because SA has a built in way of handling local optima, allowing swap to continue to improve
solution quality despite swap's much smaller local neighborhood.

\subsection{LCS Results}\label{sec:lcs}

Consider two sets of experiments with the LCS problem, one with random graphs and the other
with strongly regular graphs. In the random graph case, we have 50 instances, each consisting
of a pair of randomly generated isomorphic graphs. We use pairs of isomorphic graphs so that 
each problem instance has a known optimal solution. That is, the largest common subgraph is 
simply the graph itself. Each graph has 50 vertexes and edge density 0.5, which means that 
the probability of each possible edge is 0.5. Thus, each graph has 50 vertexes, and the 
expected number of edges is: $0.5 \cdot 50 \cdot 49 / 2 = 612.5$. The second
graph for each instance is formed by relabeling the vertexes randomly.

\begin{figure}[H]
\begin{tikzpicture} 
    \begin{scope} [vertex style/.style={draw, circle, minimum size=4mm, inner sep=0pt, outer sep=0pt, shade, top color=blue, bottom color=purple}] 
      \path \foreach \i in {0,...,24}{%
	   (14.4*\i:3.0) coordinate[vertex style] (b\i)}
       ;
	  \path \foreach \i in {0,2,4,6,8,10,12,14,16,18,20,22,24}{%
	   (14.4*\i:2.0) coordinate[vertex style] (a\i)}
       ;
	  \path \foreach \i in {1,3,5,7,9,11,13,15,17,19,21,23}{%
	   (14.4*\i:1.0) coordinate[vertex style] (a\i)}
       ; 
    \end{scope}
     \begin{scope} [edge style/.style={draw=red,double=brown}]
       \foreach \i in {0,...,24}{%
       \pgfmathtruncatemacro{\nextb}{mod(\i+1,25)}
       \pgfmathtruncatemacro{\nexta}{mod(\i+2,25)} 
       \draw[edge style] (a\i)--(b\i);
       \draw[edge style] (a\i)--(a\nexta);
       \draw[edge style] (b\i)--(b\nextb);
       }  
     \end{scope}
  \end{tikzpicture}
\caption{The generalized Petersen graph, $G(25, 2)$, a strongly regular graph.\label{fig:GenPetersen}}
\end{figure}
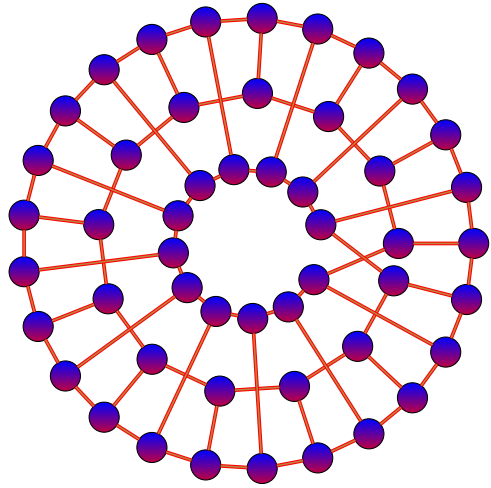

We use generalized Petersen graph $G(25,2)$ for the strongly regular case. 
Generalized Petersen graph~\cite{Watkins1969} $G(n,k)$ has $2n$ vertexes
$V = \{ u_0, u_1, \ldots, u_{n-1}, v_0, v_1, \ldots, v_{n-1} \}$, and 
$3n$ edges
$E = \{ (u_i, v_i) \mid 0 \leq i < n \} \bigcup \{ (u_i, u_{i+1 \mod n}) \mid 0 \leq i < n \} \bigcup \{ (v_i, v_{i+k \mod n}) \mid 0 \leq i < n \}$.
The original Petersen graph (Figure~\ref{fig:petersen}) is $G(5,2)$. 
Figure~\ref{fig:GenPetersen} shows generalized Petersen 
graph $G(25,2)$, which has 50 vertexes and 75 edges. We again 
average over 50 runs, but in this case each instance consists of 
graph $G(25,2)$ and a second graph isomorphic to it that is formed 
by randomly relabeling the vertexes.

\begin{figure}[H]
\begin{adjustwidth}{-\extralength}{0cm}
\begin{tabular}{cc}
\includegraphics[width=3.47 in]{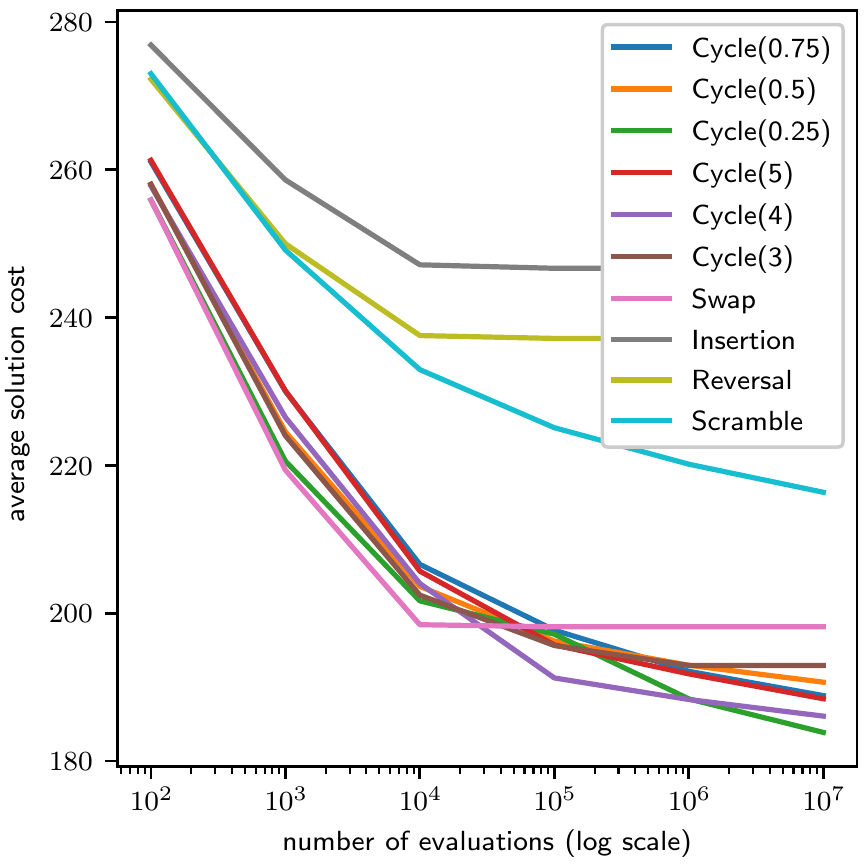} &
\includegraphics[width=3.47 in]{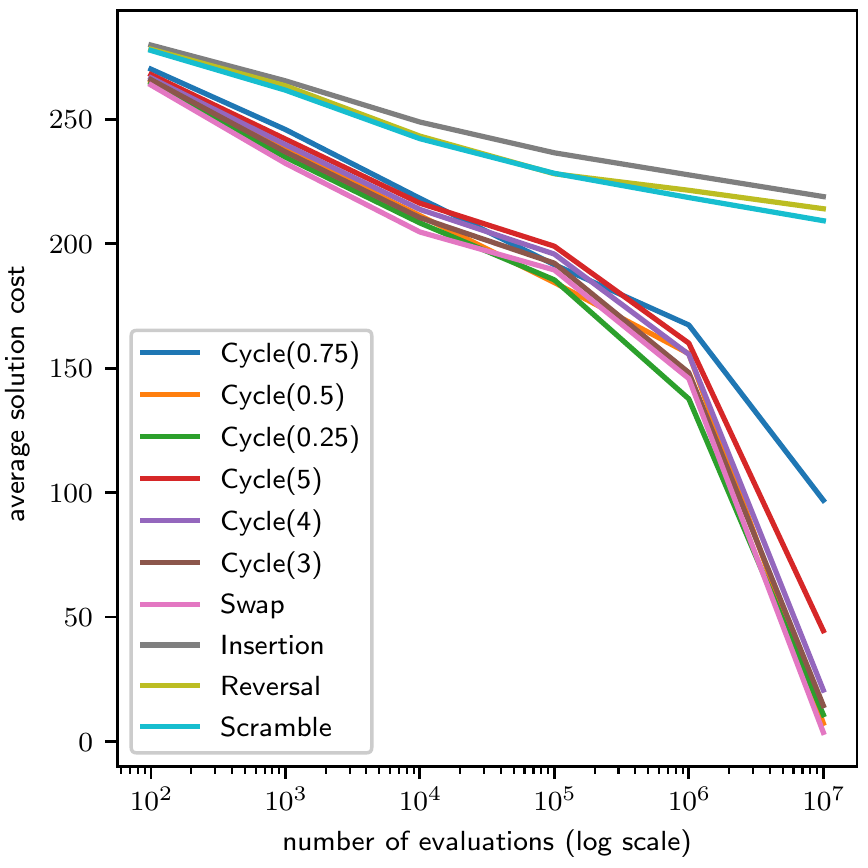} \\
(\textbf{a}) & (\textbf{b})
\end{tabular}
\end{adjustwidth}
\caption{Results on the LCS problem with random 
graphs for: (\textbf{a}) $(1+1)$-EA, and (\textbf{b}) SA.\label{fig:lcs}}
\end{figure}

LCS is a maximization problem, however, for consistency we transform
it to a minimization problem. Since each instance is a pair of isomorphic graphs, 
the LCS has $\lvert E \rvert$ edges, where $E$ is the edge set. Thus, 
redefine LCS to minimize the cost of 
permutation $p$: $C(p) = \lvert E \rvert - \lvert E^{\prime} \rvert$, where
$E^{\prime}$ is the edge set of the subgraph implied by vertex mapping $p$.

\begin{figure}[H]
\begin{adjustwidth}{-\extralength}{0cm}
\begin{tabular}{cc}
\includegraphics[width=3.47 in]{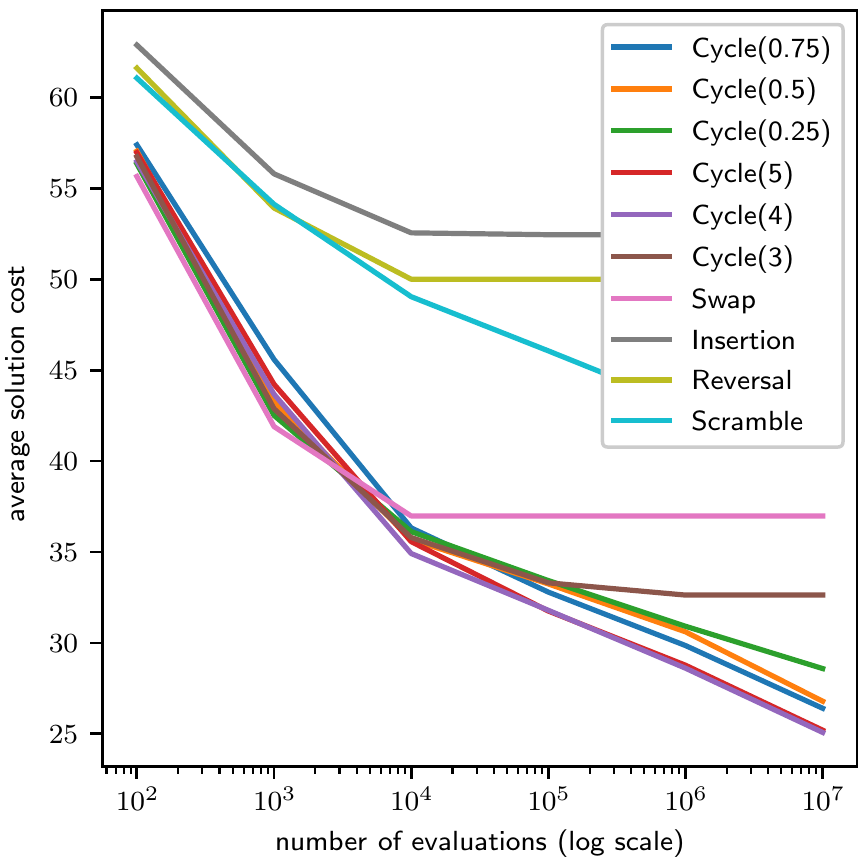} &
\includegraphics[width=3.47 in]{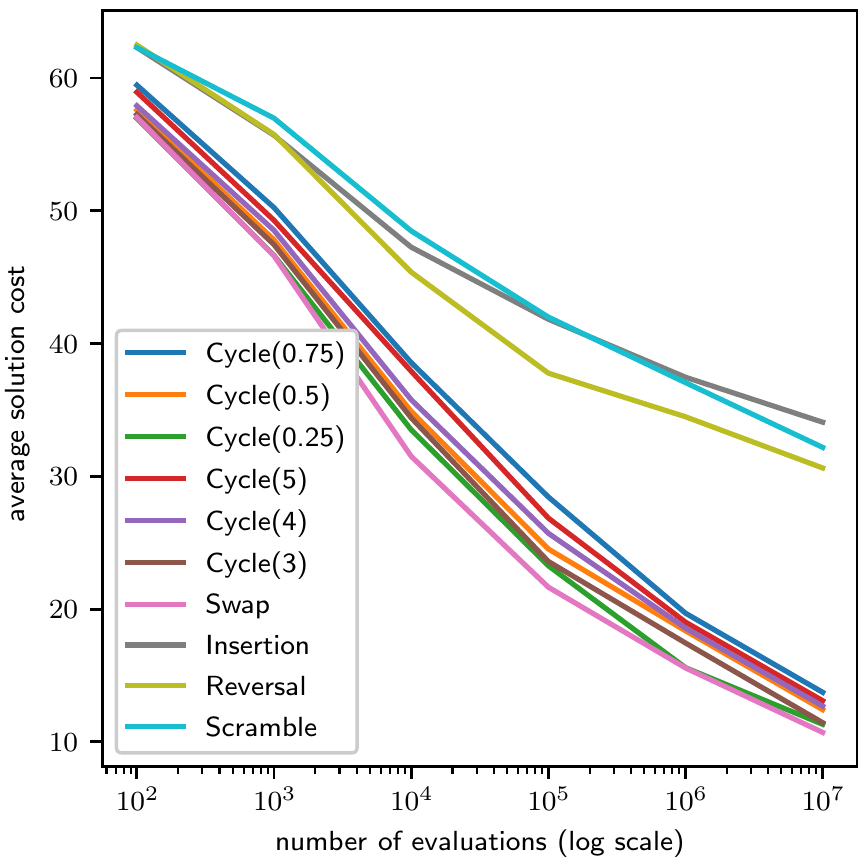} \\
(\textbf{a}) & (\textbf{b})
\end{tabular}
\end{adjustwidth}
\caption{Results on the LCS problem with strongly regular 
graphs for: (\textbf{a}) $(1+1)$-EA, and (\textbf{b}) SA.\label{fig:lcsSRG}}
\end{figure}

The results are in Figure~\ref{fig:lcs} (random graphs) and Figure~\ref{fig:lcsSRG}
(strongly regular graphs). For SA and random graphs (Figure~\ref{fig:lcs}(b)), scramble, 
insertion, and reversal all perform very poorly compared with the others. 
$\mathrm{Cycle}(5)$ and $\mathrm{Cycle}(0.75)$ are inferior to the other 
cycle mutation cases for the longest runs at statistically significant levels. Within 
an EA for random graphs (Figure~\ref{fig:lcs}(a)), the behavior is similar to that of the QAP. 
Swap exhibits a premature convergence effect, while cycle mutation finds 
increasingly lower cost solutions. $\mathrm{Cycle}(3)$ also prematurely converges
at $10^6$ evaluations, while cycle mutation configured with a larger 
neighborhood continues to improve solution cost beyond that point. 
$\mathrm{Cycle}(0.25)$ is best at statistically significant levels for the  
$10^7$ evaluation runs. 

These results are consistent with the fitness landscape analysis. First, FDC 
suggested that swap, $\mathrm{Cycle}(3)$, and $\mathrm{Cycle}(5)$ were best suited 
to the problem, followed by $\mathrm{Cycle}(4)$ and insertion. The FDC analysis 
was likely mislead with respect to insertion mutation due to the small size of the 
instance used in the FDC analysis, which is likely why we find insertion performing 
poorly on the larger instance. This is consistent with the search landscape calculus 
analysis which suggests that swap and cycle mutation are both well suited to the 
problem, and that the other operators including insertion are not. 
The $\mathrm{Cycle}(0.25)$ is most effective within longer runs of the EA 
due to the combination of small average change (average cycle length of 2.33)
and very large neighborhood size. Within SA, cycle mutation is neither better nor worse
than swap, converging at the same rate.

Observe that the strongly regular graph results (Figure~\ref{fig:lcsSRG})
are similar to those of random graphs, but are more pronounced.
For example, in an EA (Figure~\ref{fig:lcsSRG}(a)) swap and $\mathrm{Cycle}(3)$
prematurely converge; while $\mathrm{Cycle}(4)$ and $\mathrm{Cycle}(5)$ exhibit
superior convergence effect, outperforming all others at statistically significant 
levels from $10^4$ evaluations onward, but differences between them are not significant.

\section{Discussion and Conclusions}\label{sec:conclusions}

In this paper, we proposed a new mutation operator for use by EA and
other related algorithms like SA when evolving permutations. This new
operator is called cycle mutation, and includes two variations:
$\mathrm{Cycle}(\alpha)$ and $\mathrm{Cycle}(\mathit{kmax})$. Cycle
mutation induces a random permutation cycle. The difference between
the two operators is in how the cycle length is chosen. The 
$\mathrm{Cycle}(\mathit{kmax})$ operator has a small maximum cycle 
length, while $\mathrm{Cycle}(\alpha)$ does not impose any maximum
cycle length. Thus, $\mathrm{Cycle}(\alpha)$ has a significantly
larger neighborhood size, while retaining a low average cycle length.

The runtime of cycle mutation in the worst case is linear,
like insertion, reversal, and scramble mutation; but unlike those
operators, cycle mutation's average runtime is constant. Therefore,
the computational time to use cycle mutation within an EA is little more
than that of the constant time swap. Helping to achieve this efficient runtime,
cycle mutation relies on a new algorithm for sampling $k$ elements 
from an $n$ element set, called insertion sampling,
which is faster than existing alternatives for low $k$.

The fitness landscape analysis showed that cycle mutation 
is well-suited to permutation problems like the QAP and LCS, where 
absolute positions more greatly impact fitness than relative 
ordering. The fitness landscape analysis also showed that cycle 
mutation may be relevant to relative ordering problems like the TSP, 
provided cycle length is kept low. While undertaking the fitness landscape 
analysis, we developed three new measures of permutation distance: cycle 
distance, $k$-cycle distance, and cycle edit distance.

Validating the fitness landscape analysis, cycle mutation 
experimentally outperformed the others for QAP and LCS within 
an EA, especially for long runs, finding lower cost solutions 
with fewer evaluations. Furthermore, while swap suffers from a
premature convergence effect due to small neighborhood,
cycle mutation continues to make optimization progress
even for the longest runs; and the $\mathrm{Cycle}(\alpha)$ form of 
cycle mutation is especially good at managing local optima, due to 
its very large neighborhood size enabling it to make large jumps when 
necessary. Thus, $\mathrm{Cycle}(\alpha)$ exhibits superior convergence
effect.

Cycle mutation does have limitations. Cycle mutation does not
show any advantage within SA, although its rate of convergence is similar to
that of swap in SA so it may be worth considering for SA none-the-less.
Additionally, within an EA, $\mathrm{Cycle}(\mathit{kmax})$ may exhibit a
premature convergence effect for some problems if $\mathit{kmax}$ is set too
low, like we saw with $\mathrm{Cycle}(3)$ for the LCS. However, $\mathrm{Cycle}(\alpha)$
overcomes this limitation.

Our Java implementations of cycle mutation are integrated into
an open source library, Chips-n-Salsa~\cite{Cicirello2020joss}, and our Java
implementations of cycle distance, $k$-cycle distance, and cycle edit distance
are integrated into the open source JavaPermutationTools~\cite{cicirello2018} 
library. By disseminating the implementations in open source libraries, we hope 
to contribute not only to the research literature, but also to the state of 
practice. Additionally, all of the code to reproduce the experiments, and to 
analyze the results, is available in a GitHub 
repository, \url{https://github.com/cicirello/cycle-mutation-experiments}.

\vspace{6pt} 


\funding{This research received no external funding.}

\institutionalreview{Not applicable.}

\informedconsent{Not applicable.}

\dataavailability{All experiment data (raw and post-processed) is 
available on GitHub, \url{https://github.com/cicirello/cycle-mutation-experiments},
which also includes all source code of our experiments, as well
as instructions for compiling and running the experiments.} 

\conflictsofinterest{The authors declare no conflict of interest.}

\abbreviations{Abbreviations}{
The following abbreviations are used in this manuscript:\\

\noindent 
\begin{tabular}{@{}ll}
CX & Cycle Crossover \\
EA & Evolutionary Algorithm \\
ES & Evolution Strategies \\
FDC & Fitness Distance Correlation \\
GA & Genetic Algorithm \\
LCS & Largest Common Subgraph \\
QAP & Quadratic Assignment Problem \\
SA & Simulated Annealing \\
TSP & Traveling Salesperson Problem 
\end{tabular}}

\begin{adjustwidth}{-\extralength}{0cm}

\reftitle{References}


\bibliography{paper.bib}

\end{adjustwidth}
\end{document}